\documentclass{article}
\pdfoutput=1

\usepackage[nonatbib, preprint]{neurips_2021}

\usepackage[utf8]{inputenc} 
\usepackage[T1]{fontenc}    
\usepackage{hyperref}       
\usepackage{url}            
\usepackage{booktabs}       
\usepackage{amsfonts}       
\usepackage{nicefrac}       
\usepackage{microtype}      
\usepackage{xcolor}         

\usepackage{amsmath,amssymb}
\usepackage{mathtools}
\usepackage{amsthm}
\usepackage{tikz}
\usetikzlibrary{arrows.meta,backgrounds,calc}
\usepackage[colorinlistoftodos]{todonotes}
\usepackage{enumitem}
\usepackage{graphicx}
\usepackage{graphbox}
\usepackage{booktabs} 
\usepackage{caption}
\usepackage[subrefformat=parens]{subcaption}
\usepackage{hyphenat}
\usepackage{calc}
\usepackage{adjustbox}

\theoremstyle{definition}
\newtheorem*{notation}{Notation}

\newcommand{\flexparen}[1]{\left(#1\right)}
\newcommand{\set}[1]{\{#1\}}
\newcommand{\flexset}[1]{\left\{#1\right\}}

\newcommand{\cmb}{,\linebreak[0]}  

\makeatletter
\newcommand{\mask}[2]{{\mathpalette\mask@{{#1}{#2}}}}
\newcommand{\mask@}[2]{\mask@@{#1}#2}
\newcommand{\mask@@}[3]{%
  \settowidth{\dimen@}{$\m@th#1#2$}%
  \makebox[\dimen@]{$\m@th#1#3$}%
}
\makeatother

\makeatletter
\let\c@figure\c@table
\let\ftype@figure\ftype@table
\makeatother

\title{When adversarial examples are excusable}

\author{%
  Pieter-Jan Kindermans\thanks{Both authors contributed equally}\\
  Google Research\\
  \texttt{pikinder@google.com} \\
  \And
  Charles Staats$^{*}$ \\
  Google Research \\
  \texttt{cstaats@google.com}
}

\begin{document}

\maketitle

\begin{abstract}
Neural networks work remarkably well in practice and theoretically they can be universal approximators.
However, they still make mistakes and a specific type of them called adversarial errors seem inexcusable to humans. 
In this work, we analyze both test errors and adversarial errors on a well controlled but highly non-linear visual classification problem. 
We find that, when approximating training on infinite data, test errors tend to be close to the ground truth decision boundary. Qualitatively speaking these are also more difficult for a human. By contrast, adversarial examples can be found almost everywhere and are often obvious mistakes.
However, when we constrain adversarial examples to the manifold, we observe a 90\% reduction in adversarial errors.
If we inflate the manifold by training with Gaussian noise we observe a similar effect.  In both cases, the remaining adversarial errors tend to be close to the ground truth decision boundary. Qualitatively, the remaining adversarial errors are similar to test errors on difficult examples. They do not have the customary quality of being inexcusable mistakes. 
\end{abstract}

\section{Introduction}
Adversarial examples are tiny, imperceptible perturbations that transform an initially correctly classified image into a wrongly classified one \cite{szegedy2014intriguing}. This effect is most often illustrated with images that end up being obviously misclassified by a machine learning model, but are perceived by a human as being unmodified \cite{goodfellow2014explaining}. Many people have proposed defenses against adversarial attacks \cite{roth2019odds,cisse2017parseval,song2017pixeldefend,samangouei2018defense,buckman2018thermometer}, but more often than not, a defense can be circumvented \cite{athalye2018obfuscated,DBLP:conf/iclr/MadryMSTV18,goodfellow2014explaining,carlini2017adversarial,DBLP:conf/iclr/KurakinGB17a,dong2019evading,li2019nattack}. Currently there is not a real consensus on what exactly causes these adversarial examples. 

Initially Goodfellow et al. \cite{goodfellow2014explaining} attributed the existing of adversarial examples due to the models being too linear. In essence, having high dimensional dot products makes it easy to manipulate the prediction. 
Recent work \cite{DBLP:conf/nips/IlyasSTETM19} argues that the directions in which the classifier can be manipulated are not necessarily erroneous. They show that typical image datasets contain predictive features that cannot be perceived by humans. The classifier can react to these features and this causes the adversarial perturbations.
A third perspective can be found in the work by Gilmer et al.\footnote{An earlier, widely cited version of that paper had Nic Ford as a first author.} \cite{DBLP:conf/icml/GilmerFCC19} which argues that if the model performs poorly on images corrupted with Gaussian noise, it should not be surprising that we can find mistakes for the classifier close to the original datapoint.

There is one more hypothesis that received remarkably little attention: the off manifold hypothesis. In statistical learning theory \cite{vapnik2013nature} it is generally assumed that the training and the testing data come from the same distribution. Since adversarial examples have much higher
error rates, they clearly are drawn from a different distribution. But since ``out-of-distribution'' is a notion
that applies poorly to individual images, we instead ask whether
the data still lies on the data manifold after it is modified by an adversarial perturbation. Deciding whether a data point is on or off manifold is incredibly hard for natural images. For example, different lenses introduce different image corruptions, and compression can also introduce artifacts. Furthermore defenses built on the off-manifold hypothesis \cite{song2017pixeldefend, samangouei2018defense} have a poor record of success \cite{athalye2018obfuscated}. Taking another approach, Stutz et al. \cite{Stutz_2019_CVPR} used VAE-GANs to find adversarial examples on a learned data manifold. Because imperfect manifold learning was a confounding factor, they performed an additional experiment using data from a known generative model. In this experiment, they found that most adversarial examples left the data manifold and on manifold robustness seemed to correlate with generalization. In this paper we push these results further.

To be able to analyze adversarial examples better we cannot just rely on  natural image datasets e.g. ImageNet~\cite{russakovsky2015imagenet} and CIFAR-10~\cite{krizhevsky2009learning}.
While one could argue represent real world use cases with imperfectly controlled data. For example, a key limitation of ImageNet is the simple fact that there are images for which the ground truth label is not uniquely defined; one image prominently features both a dog and a pile of cherries \cite{alexnet}. Additionally, given that we do not fully understand the generative model of the data, we cannot say with certainty whether a certain perturbation is in or out of distribution. Even when attempts are made at creating a new test set for ImageNet, there is evidence that the distribution has changed \cite{DBLP:conf/icml/RechtRSS19}. The lack of well controlled data prevents researchers from 
performing precise experiments.

Because investigating the manifold hypothesis is so difficult on natural image datasets we propose the Squiggles family of datasets. These comprise images of self intersecting and non-intersecting curves. This classification task is high-dimensional but interpretable to a human, unlike e.g. the adversarial spheres data \cite{DBLP:conf/iclr/GilmerMFSRWG18}, which humans cannot solve in high dimensions. Furthermore, the generative process is fully controlled and differentiable. This allows us to backpropagate all the way to the latent space. On top of that, the correct label for any point on the data manifold can be computed automatically, enabling the use of effectively unlimited data.
By navigating the data manifold and recomputing the true label, we can estimate the distance between the true decision boundary and any sample on the data manifold. This gives us a measure of the true difficulty of a sample. Since we control the generative process we can make the distinction between on- and off-manifold samples.

We make use of these aspects in our study on adversarial examples. 
We show that when a model is trained on a large number of samples (66 million) most generalization errors are close to the decision boundary, i.e., objectively difficult. Ordinary adversarial examples are still common in this setting. But because we control the generative process we can say with certainty that they do not belong to the data manifold, a point which is open for debate on natural images.
When we control for going off manifold, either by incorporating small perturbations in the training data or by restricting the search for adversarial examples to the data manifold, we observe an over 90\% decrease in the effectiveness of the adversarial attack. Furthermore, we provide evidence that adversarial attacks are only successful on a limited subset of samples in this case. These ``vulnerable" samples are more difficult to classify for a human and are also closer to the true decision boundary.
In essence these results show that with a very large amount of accurately labelled data, it is technically possible to build a classifier that behaves as a universal approximator and does not make inexcusable mistakes on limited size perturbations. We do show as well that as dimensionality increases, achieving this becomes even more difficult in practice.  

\begin{figure}[b]
\centering
 \begin{subfigure}{0.3\columnwidth}
 \includegraphics[width=\textwidth]{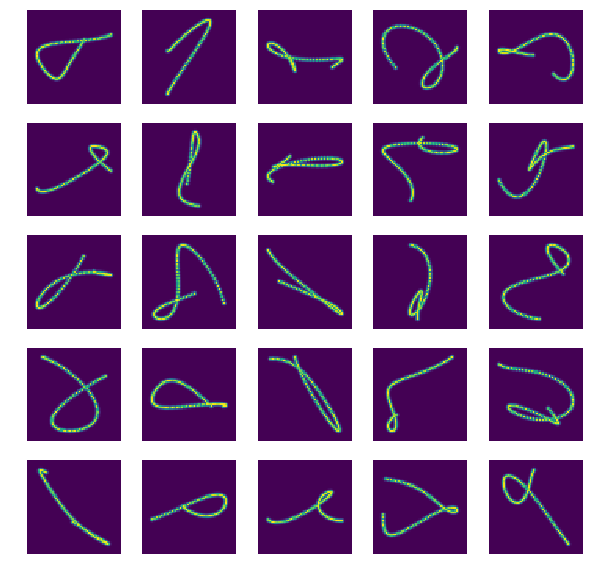}
 \caption{self-intersecting}\label{fig:rnd_self}
 \end{subfigure}
  \begin{subfigure}{0.3\columnwidth}
 \includegraphics[width=\textwidth]{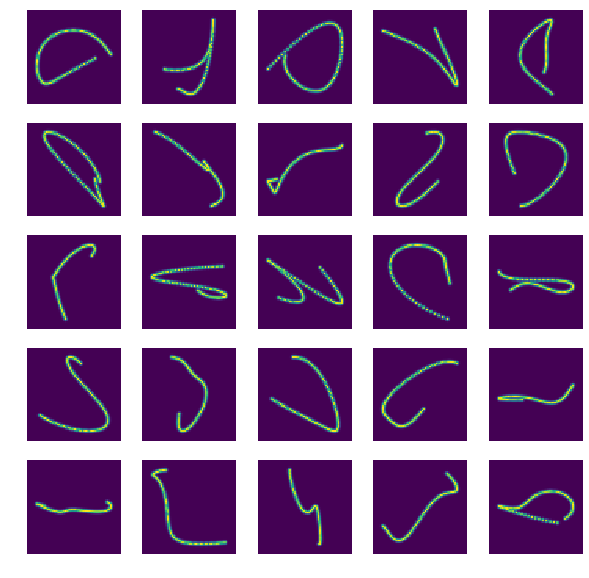}
  \caption{not-intersecting}
 \end{subfigure}
 \caption{The SineNet dataset. Twenty-five random examples from each class are shown.}\label{fig:dataset}
\end{figure}

\begin{figure*}
\begin{adjustbox}{width=\linewidth}
\begin{tikzpicture}[data/.style={
                      rectangle,
                      draw=black,
                      fill=white,
                      node font=\large
                    },
                    arrow label/.style={
                      text depth=1.5ex,
                      text height=3.0ex,
                      node font=\small
                    }
                   ]
  \matrix[row sep=0mm,column sep=7mm] {
    \node (z) [data] {$\mathbf{z}$}; &
    \node (z2c) [arrow label,above] {\parbox{\widthof{coordinates}}{\raggedright{}Latent to coordinates}}; &
    \node (c) [data] {$\mathbf{c} = {\small
      \begin{pmatrix*}[r]
        0.1 & 0.2 \\
        0.2 & 0.2 \\
        0.4 & 0.6 \\
        \mathrlap{\;\vdots} \\
        0.9 & 0.5
      \end{pmatrix*}
      }$}; &
    \node (c2x) [arrow label,above] {\parbox{\widthof{Coordinates}}{Coordinates to \smash{pixels}}}; &
    \node (x) [data] {$\mathbf{X}={}$\includegraphics[width=1cm,height=1cm,align=c]{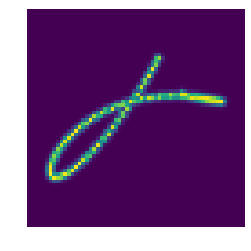}}; &
    \node (x2yhat) [arrow label,above] {\parbox{\widthof{classifier}}{\raggedright{}Image classifier}}; &
    \node (yhat) [data] {$\hat{y}=0.76$};
    \\
    & & & &
    \node (c2y) [arrow label] {\parbox{\widthof{Coordinates to label}}{Coordinates to label \phantom{x}}}; & &
    \node (y) [data] {$y=1$}; \\
  };
  \begin{scope}[on background layer,
                every path/.style={line width=1pt,shorten >=2pt}]
    \draw[->] (z.mid east) -- (c.mid west);
    \draw[->] (c.mid east) -- (x.mid west);
    \draw[->] (x.mid east) -- (yhat.mid west);
    \coordinate (sinstart) at ($0.5*(c.south east) + 0.5*(c.mid east)$);
    \draw[->] (sinstart) cos ($0.5*(sinstart) + 0.5*(c2y.mid west)$) sin (c2y.mid west) -- (y);
  \end{scope}
\end{tikzpicture}
\end{adjustbox}
\caption{The entire process from latent space to classification is made up of differentiable components. 
First, we start in latent space and compute the coordinates for 100 points on the curves. These coordinates are used to compute the label and can be transformed into an image. Afterwards a standard image classifier can be applied.}\label{fig:generative}
\end{figure*}
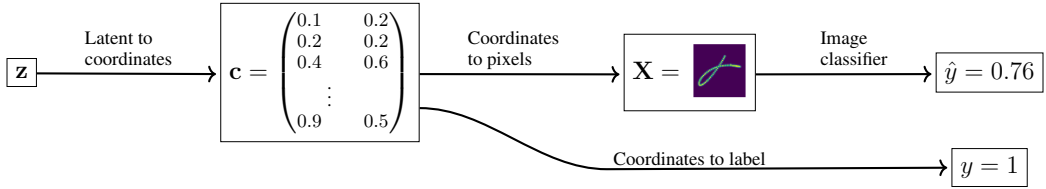

\section{The Squiggles dataset}
The analysis we want to do in this paper cannot be done on  natural images because we cannot obtain completely accurate labels or infinite data. Therefore we propose the Squiggles dataset\footnote{Code to generate the dataset is available at https://github.com/google-research/google-research/tree/master/squiggles},
which we use at resolution $60\times60$. 

The curves in the squiggles dataset are generated by a process illustrated in Fig.~\ref{fig:generative}. We start with a latent representation $\mathbf{z}$ which parametrizes a function $\boldsymbol{\Theta}_{\mathbf{z}}(t)$ that is able to generate a curve.
We have multiple possible parametrizations. For the SineNet variant, this latent representation specifies the weights in a neural network with cosine activations. In the Taylor variant the parameters are the coefficients of a Taylor approximation of a curve. Given a sample of latent variables, we can compute 100 coordinates $\mathbf{c}$ on the curve by applying $\boldsymbol{\Theta}_{\mathbf{z}}(t)$ to fixed inputs $t$. 
The coordinates $\mathbf{c}$ are used to compute the label and are also transformed into an image $\mathbf{x}$. The only component that is not continuous and piecewise differentiable is the computation of the label. See Appendix \ref{app:squiggles} for a detailed explanation of the generative process.

\section{Distance to the decision boundary as a measure of difficulty}\label{sec:distance_to_bdry}
The fully differentiable generative process allows us to navigate the data manifold. This allows us to estimate the distance to the ground truth decision boundary. The distance to the ground truth decision boundary functions as a model-independent measure of difficulty.

\begin{figure*}
        \centering
\begin{subfigure}{\textwidth}
    \includegraphics[width=\textwidth]{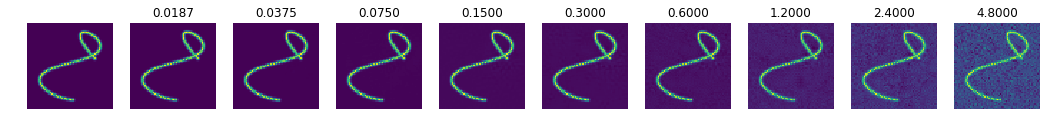}
    \caption{pixel space}
    \label{fig:pgd_pixel}
\end{subfigure}

\begin{subfigure}{\textwidth}
\includegraphics[width=\textwidth]{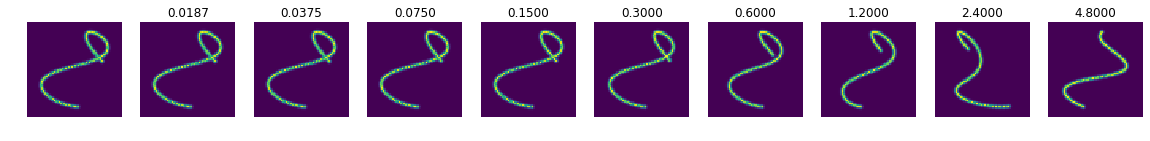}
    \vspace{-0.7cm}
    \caption{latent space}
    \label{fig:pgd_latent}
\end{subfigure}
\caption{The effect of an $\ell^2$ PGD attack in \subref{fig:pgd_pixel} pixel space and \subref{fig:pgd_latent} latent space for different values of $\epsilon$. In contrast to a pixel space attack, an attack in latent space can be semantically meaningful.}
\end{figure*}

We measure this distance in latent space since the ground truth decision boundary is not defined in pixel space. Note that by continuity, a sufficiently small change in latent space results in a small change in pixel space; so we can guarantee that when two examples are close together in latent space they are visually similar. See also Fig.~\ref{fig:pgd_latent} to observe small latent-space changes in practice.

To estimate an upper bound on the distance to the ground truth decision boundary we perform a targeted adversarial attack in latent space. For this we use a PGD attack \cite{DBLP:conf/iclr/MadryMSTV18} since the model's gradient w.r.t. the latent variables provides an approximate direction toward the decision boundary.  While the direction is approximate, we do recompute the ground truth label to verify whether this crossed the true decision boundary or not. This gives us the upper bound on the distance to the ground truth decision boundary. To reduce the likelihood that this approximate direction is subject to the same errors as the model we are analyzing, we \emph{intentionally} use a model trained on the Taylor variation of the data to navigate the latent space when we are analyzing SineNet models.

To approach the decision boundary along the manifold we use a PGD attach with $\ell^2$ norm $\epsilon$, $100$ iterations and the stepsize per iteration $\epsilon_i=\epsilon/25$. 
We perform the attack using nine different logarithmically spaced values for $\epsilon$ ranging from 0.01875 to 4.8. The estimate of the distance to the decision boundary is the smallest $\epsilon$ for which the ground truth label (which we can compute automatically) changes. The effect of the attack at different magnitudes is visualized in Fig.~\ref{fig:pgd_latent}. 

\begin{table}
 \centering{\small
 \begin{tabular}{lrr}
 \toprule
      & non-inters & self-inters\\
    \midrule

      {\bf Taylor} &  88.4\% & 21.6\% \\
      {\bf SineNet} & 65.2\% & 34.8\% \\
      \bottomrule \\
 \end{tabular}}
 \caption{Label distribution for Taylor and SineNet.\label{table:label_distributions}}
\end{table}

\section{Experiments}
\subsection{Setup}
In this work, we focus our analysis on a ResNet-34 \cite{he2016deep}. The ResNet was chosen simply because it is a commonly used architecture.

We train the network for 129600 steps with a batch size of 512 without ever repeating samples. Thus, training dataset consists of 66,355,200 ($=2^{26}$) samples. This is about 55 times larger than the ImageNet training set. We wanted to investigate classifiers approaching the limit of infinite data.

The test set size is more modest with 98,304 samples (3 shards of 32768 samples), which is still almost twice the size of the ImageNet validation set. While it is technically possible to generate the data on-line, we found that training time was dominated by in-process labeling, so we generated distinct train and test sets prior to performing our experiments. Note that we prioritized ease-of-use over efficiency when choosing our labeling library, so this decision could be revisited for sufficient cause.

Table~\ref{table:hyper} gives an overview of the training hyperparameters. The values for the hyperparameters were selected as they enabled us to train a high quality model (0.999 roc-auc) on the Taylor data in about 9 hours on a single GPU. After the hyperparameters were set, new datasets were generated to perform the actual experiments.

The CleverHans library \cite{papernot2018cleverhans} was used for all adversarial attacks.

\begin{table}
    \centering{\small
 \begin{tabular}{ll}
 \toprule
     {\bf Hyperparameter} & Value\\
    \midrule
    Training set size & 66,355,200 \\
    Test set size & 98,304 \\
    Training Hardware & 1 GPU\\
    Batch size & 512\\
    Training steps & 129600 \\
    Learning rate decay schedule & Cosine Decay \\
    Initial learning rate & 4e-4 \\
    Optimizer & Adam $\beta_1=0.9$, $\beta_2=0.999$,$\epsilon=10^{-8}$\\
    \bottomrule \\
 \end{tabular}}
 \caption{Hyperparameters for the basic experiments\label{table:hyper}}
 \end{table}

\subsection{Within and Cross Distribution Generalization}
The main goal of this section is to establish that the classifiers studied in this work are highly accurate and capable of generalizing to other data distributions. 

\begin{table}
    \centering
    {\small
 \begin{tabular}{ccccc}
 \toprule
      {\bf Training data} & \multicolumn{2}{c}{{\bf Taylor}} & \multicolumn{2}{c}{{\bf SineNet}}  \\
      & acc. & roc-auc. & acc. & roc-auc. \\
    \midrule

      within-distribution   & 98.95\%   & 0.9994 &   98.75\% &  0.9992 \\
      cross-distribution    & 97.59\%   & 0.9962 &   98.72\% &  0.9989 \\
      \bottomrule \\
 \end{tabular}
 }
 \caption{Within and cross-distribution evaluation.\label{table:eval}}
 \end{table}

We evaluate the accuracy and roc-auc for within and cross distribution generalization (see Table~\ref{table:eval}). For within-distribution generalization we evaluate a model trained on Taylor (SineNet) data, on the Taylor (respectively SineNet) test set. 
Within distribution, the Taylor model achieves 98.95\% test accuracy and the SineNet model has 98.75\% test accuracy.

Based on these results, one could mistakenly assume that this problem is quite trivial, but it is not. It is highly non-linear and benefits from the vastness of our training sets. The same training setup (SineNet within-distribution) but limited to 32758 samples (allowing multiple epochs to reach the same overall number of training steps) only achieves 89.35\% accuracy. If we increase this to about 2 million samples we end up at 97.95\% accuracy. This is twice the amount of data compared to ImageNet but much worse than the 66M sample result.  This shows that this problem benefits from even more data.

Given sufficient data, the  models generalizes outside the training distribution as well\footnote{We verified that the distributions from Taylor and SineNet samples are different by training a classifier that learned the differences between the distributions.}. We evaluate the Taylor model on SineNet data and vice versa.  This comes down to about 1000 mistakes on the test set. Both achieve a near-perfect roc-auc of 0.999. We hypothesise that this might be due to the SineNet dataset containing more varied or complex curves. In what follows we will focus on the SineNet model for analysis.

\subsection{Are misclassified examples objectively difficult?}
\begin{figure*}
     \centering
     \begin{subfigure}[t]{0.24\textwidth}
		\centering		
		\includegraphics[width=\textwidth]{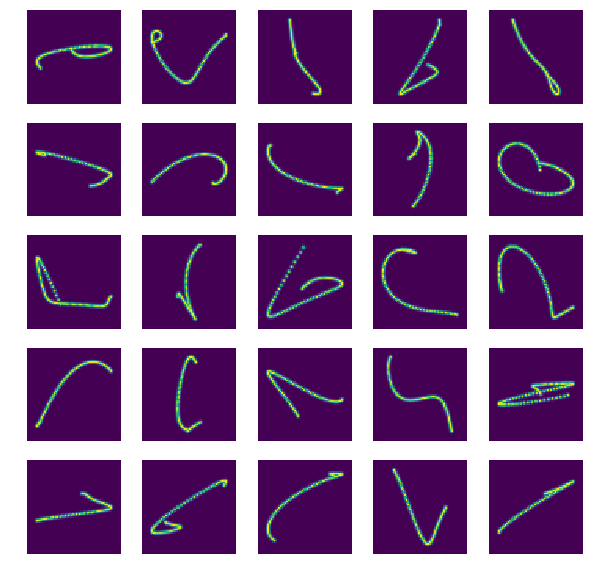}
		\caption{mistakes}\label{fig:mistake_self}		
	\end{subfigure}
    \begin{subfigure}[t]{0.24\textwidth}
		\centering		
		\includegraphics[width=\textwidth]{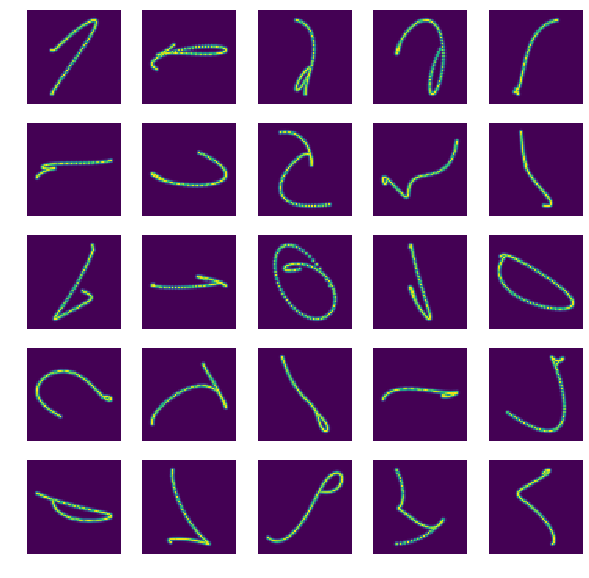}
		\caption{\raggedright adv. errors in latent space} \label{fig:adv_latent_self}
	\end{subfigure}
    \begin{subfigure}[t]{0.24\textwidth}
		\centering		
		\includegraphics[width=\textwidth]{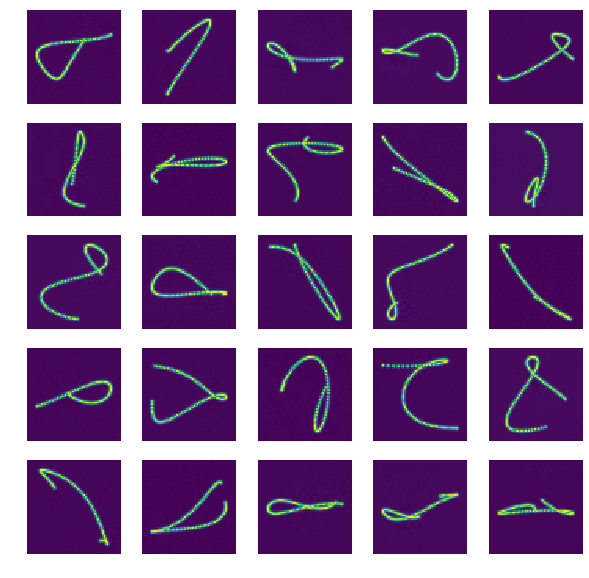}
		\caption{adv. errors ($\sigma=0$)}\label{fig:adv_self}	
	\end{subfigure}
    \begin{subfigure}[t]{0.24\textwidth}
		\centering		
		\includegraphics[width=\textwidth]{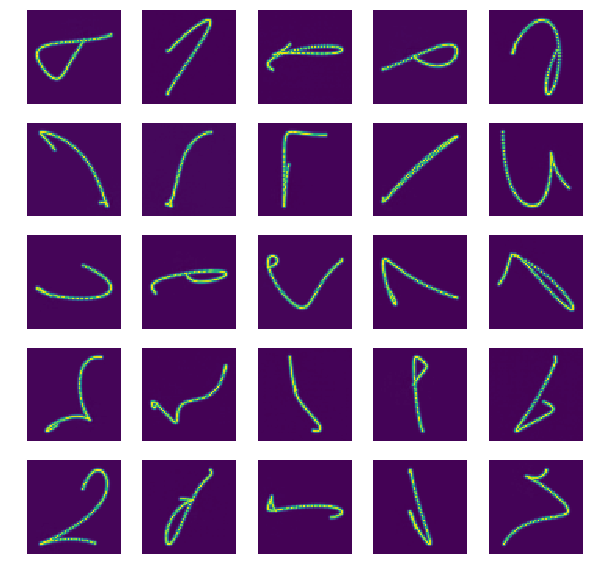}
		\caption{adv. errors ($\sigma=0.08$)}\label{fig:adv_nois_self}
	\end{subfigure}

\caption{Different types of errors made on self-intersecting samples. All examples are misclassified as non-intersecting. 
Panel~\subref{fig:mistake_self} contains generalization errors. In many cases, the self intersection is hard to spot. Panel~\subref{fig:adv_latent_self} 
shows errors made by perturbing correctly classified samples in latent space to flip the model's
prediction but not the true label. These examples appear ambiguous to humans.
Panel~\subref{fig:adv_self} contains adversarial errors for a standard model. The self intersection is often more obvious than for generalization errors. Panel~\subref{fig:adv_nois_self} shows adversarial errors for a model trained with Gaussian noise on the pixels. It has increased adversarial robustness and the mistakes are less obvious.}\label{fig:errors}
\end{figure*}

The classifiers we trained have high but not perfect accuracy. Fig.~\ref{fig:rnd_self} in the introduction showed 25 randomly selected examples of self-intersecting curves from the dataset. In most cases these samples have a big loop and a very obvious self intersection. Fig.~\ref{fig:mistake_self} gives twenty-five randomly selected examples of self-intersecting curves that were classified incorrectly as non-intersecting. In many cases, the self intersection is much harder to spot. This indicates that these examples might be more difficult. 

\begin{figure*}[t]
    \centering
    \begin{subfigure}{0.32\columnwidth}
    \includegraphics[width=\columnwidth]{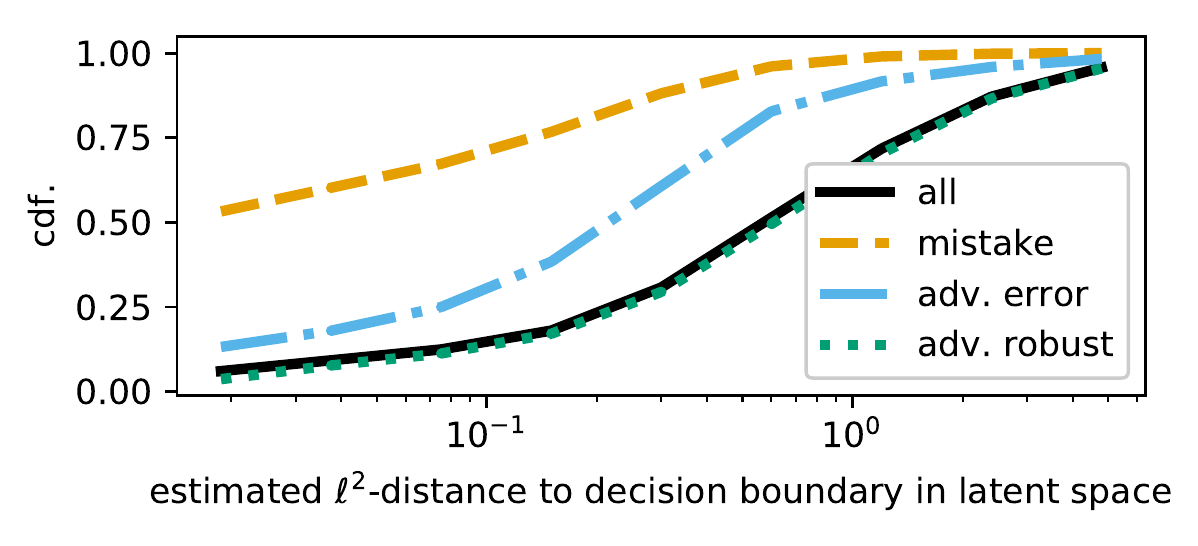}
    \caption{}
    \label{fig:distance_latent_latent_attack}
    \end{subfigure}
    \begin{subfigure}{0.32\columnwidth}
    \includegraphics[width=\columnwidth]{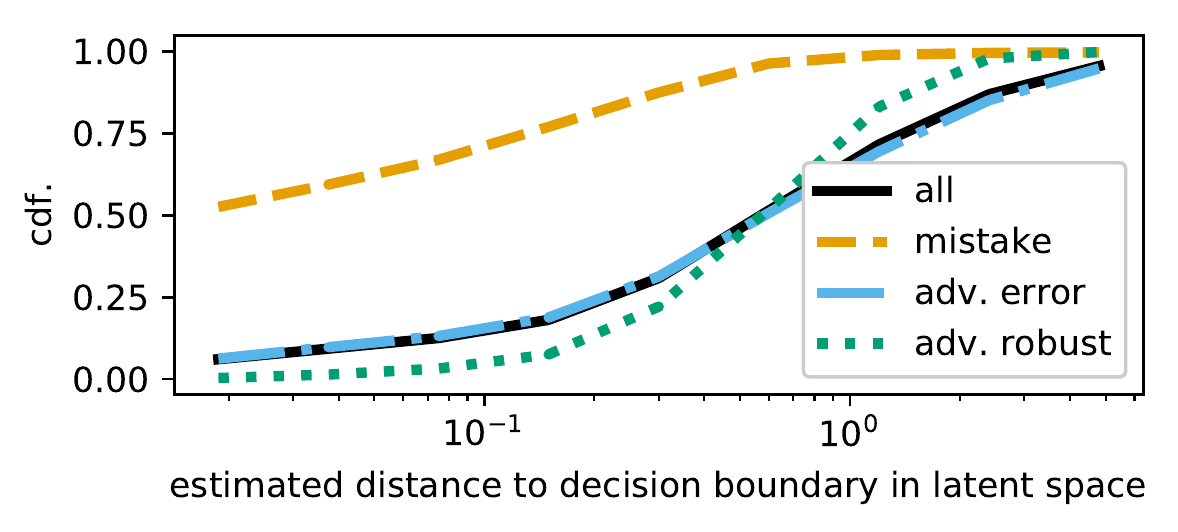}
    \caption{}
    \label{fig:distance_latent_pixel_attack}
    \end{subfigure}
    \begin{subfigure}{0.32\columnwidth}
    \includegraphics[width=\columnwidth]{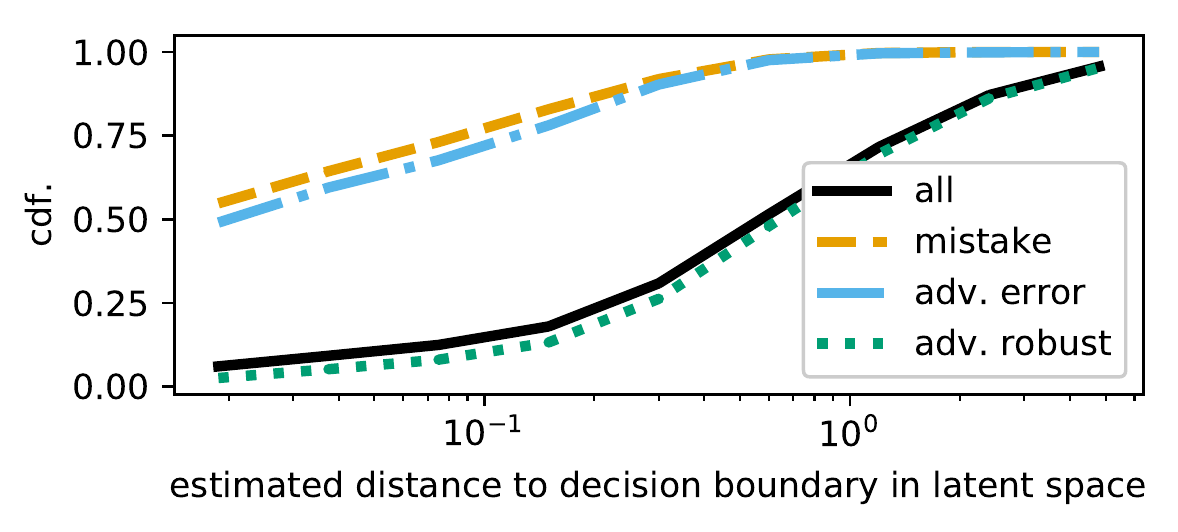}
    \caption{}
    \label{fig:distance_latent_noise}
    \end{subfigure}
    \caption{The fraction ($y$-axis) of samples within a specified distance ($x$-axis)
    of the true decision boundary. The $x$-axis is more precisely our best upper bound
    on the $\ell^2$ distance in latent space to the ground-truth decision boundary.
    \subref{fig:distance_latent_latent_attack} A vanilla model subject to attacks in latent space. The latent space attack had $\ell^2$-norm equal to 0.075, which was the most successful setting.
    \subref{fig:distance_latent_pixel_attack} The same vanilla model subject to attacks
    in pixel space.
    \subref{fig:distance_latent_noise} Pixel space attacks for a model trained with pixel-space noise $\sigma=0.08$.\hfill\vspace{0.5ex}\linebreak[5]
    For all these models, half of the regular misclassifications (dashed orange line) are at a distance of at most $0.01875$ (leftmost $x$-value) from the decision boundary.
    When the attack is in pixel space, almost any sample can be attacked regardless of
    its difficulty in a vanilla model. Correspondingly,
    the distance distribution of adversarial errors
    of a vanilla model \subref{fig:distance_latent_pixel_attack} mimics that of general
    samples. But the adversarial error distribution mimics generalization errors when the model is trained with pixel
    noise \subref{fig:distance_latent_noise}. Thus, for the noise-trained model,
    adversarial errors are almost as excusable as conventional generalization errors.}
\end{figure*}

To quantify the difficulty of the misclassified examples, we estimate the distance to the ground truth decision boundary in the way we described in Section \ref{sec:distance_to_bdry}.
We present the results in Fig.~\ref{fig:distance_latent_latent_attack}. On average, 6\% of the regular samples are at an estimated $\ell^2$ distance less than $0.01875$ from the decision boundary in latent space. For misclassified samples this rises to 53\%.  Because such a small change in latent space has a small perceptual effect, we consider these misclassified examples objectively more difficult.


\begin{figure}
    \centering
    \begin{subfigure}{0.45\columnwidth}
    \includegraphics[width=\textwidth]{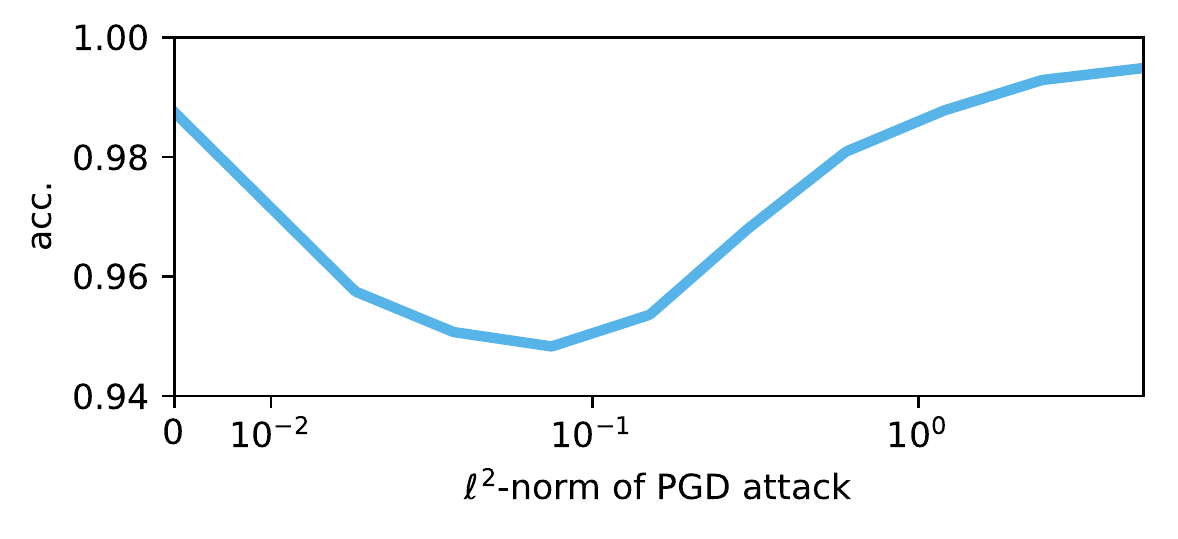}
    \end{subfigure}
    \begin{subfigure}{0.45\columnwidth}
    \includegraphics[width=\textwidth]{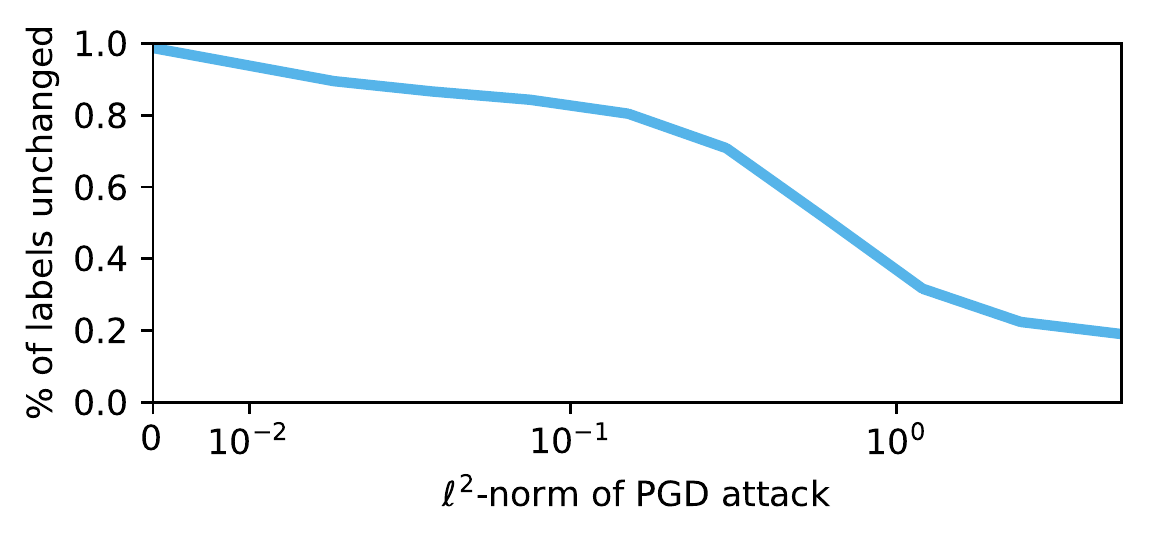}
    \end{subfigure}
    \caption{Behavior of the model under increasingly large on-manifold adversarial attacks.
     Left: Model accuracy against true label (recomputed after the attack).  Right: Fraction
     of samples whose ground-truth labels do not change after the latent-space
     attack.\hfill\vspace{0.5ex}\linebreak[5]
     Contrary to what is observed in pixel space the effectiveness of an attack does not increase monotonically with the size of the attack.
     As the attack size increases, more ground truth labels are changed but the examples also become easier.}\label{fig:eval_latent}
\end{figure}

\subsection{Can we find on-manifold adversarial errors?}
According to the original paper \cite{szegedy2014intriguing}, adversarial examples are examples that are misclassified after an imperceptible perturbation. They also argue that because the perturbation is so small, it cannot be semantically meaningful. To be able to make the argument that adversarial examples are (partially) caused by going off manifold, it should be hard to find adversarial examples on manifold. Also, to be truly adversarial we need to test whether the mistakes are blatant or subtle. 

The fully  differentiable generative process allows us to look for adversarial examples in latent space.
The adversarial attack we use is the highly effective PGD attack \cite{DBLP:conf/iclr/MadryMSTV18}.
We use an $\ell^2$-norm of $\epsilon\in\set{0.01875\cmb 0.0375\cmb 0.075\cmb 0.15\cmb 0.3\cmb 0.6\cmb 1.2\cmb 2.4\cmb 4.8}$, $100$ iterations and the stepsize per iteration $\epsilon_i=\epsilon/25$. 

In most adversarial examples research, it is assumed that small perturbations cannot change
the true label. Showcasing one advantage of the Squiggles dataset, we remove this assumption 
by recomputing the true label after perturbing the latent space coordinates.

The adversarial accuracy is given in Fig.~\ref{fig:eval_latent}. When we select the most effective attack, which has $\epsilon=0.075$, the accuracy is still $94.8\%$. By restricting ourselves to the latent space we observe only a limited success for the attack. What is striking, however, is that the accuracy does not monotonically decrease as the magnitude of the attack increases, contrary to what is commonly observed in pixel space attacks. We observe that, as shown in Fig.~\ref{fig:eval_latent}, that when the magnitude of the attack increases, we tend to flip the ground truth label more often. However, as can be seen in Fig.~\ref{fig:pgd_latent}, when we move sufficiently far in latent space, the sample might become easier to classify due to the attack. Fig.~\ref{fig:attacks_latent} shows the actual attack vectors, visualized in pixel space. With these attacks, the appearance of the curve remains largely stable. 

When we look at the distances from the decision boundary in Fig.~\ref{fig:distance_latent_latent_attack}, we observe that the adversarial errors are on average closer to the decision boundary than typical samples, despite being farther from the decision boundary than typical generalization errors.  Finally, Fig.~\ref{fig:adv_latent_self} confirms that the curves with adversarial errors also tend to be more difficult than the typical samples of Fig \ref{fig:rnd_self}. 
In this case one could argue that these adversarial errors tend to be somewhat excusable (or not that adversarial).

 \begin{figure}[t]
     \centering
     \begin{subfigure}{0.45\columnwidth}
     \includegraphics[width=\columnwidth]{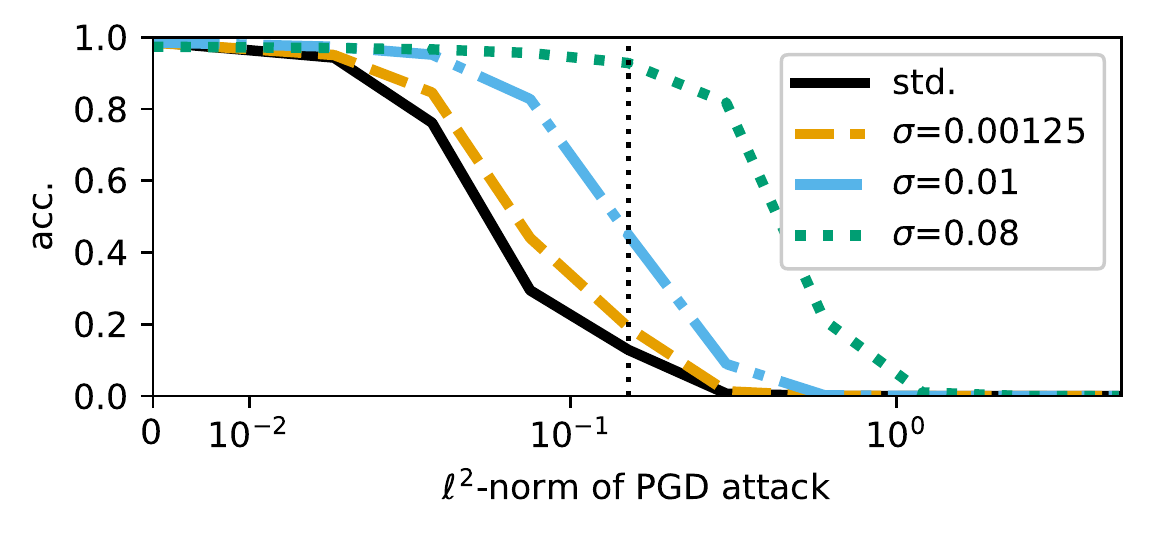}
     \end{subfigure}
     \begin{subfigure}{0.45\columnwidth}
     \includegraphics[width=\columnwidth]{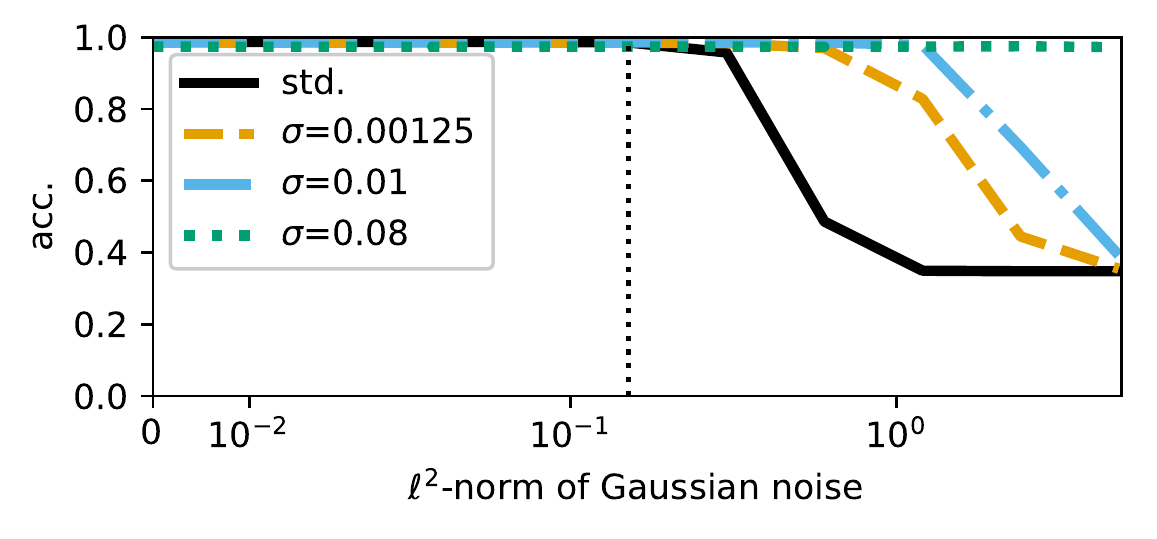}
     \end{subfigure}
     \caption{Comparison of the effectiveness of a PGD attack and a Gaussian noise attack in pixel space for models trained with different amounts of pixel noise. The dotted vertical line corresponds to the $\ell^2$-norm $0.15$, which is the focus of the analysis below.}
     \label{fig:pixel_eval}
 \end{figure}

\subsection{We can of course find adversarial perturbations in pixel space}
So far we have established that the network generalizes well within-distribution and cross-distribution.  We have also shown that the many of the mistakes the model makes within distribution are forgivable.
Even an adversarial attack restricted to the data manifold has limited effect. 
Now we turn our attention to adversarial examples in pixel space. An example of an image that is adversarially attacked at different magnitudes is given in Fig.~\ref{fig:pgd_pixel}. Below $\epsilon=0.15$, the attack is very hard to spot, above $\epsilon=0.30$ it becomes rather obvious. This makes these values a good choice for further analysis. 

The accuracy of the models after these attacks is is shown in Fig.~\ref{fig:pixel_eval}. At $\epsilon=0.15$ the accuracy drops from 98.75\% to 12.86\%. Since most examples can be attacked it should not be surprising that  we also observe that unlike generalization errors the adversarial errors are not close to the true decision boundary in latent space. (See Fig.~\ref{fig:distance_latent_pixel_attack}.)

Why does a network that has good within and cross-distribution generalization have so many adversarial examples? Because we control the generative process, we can state with high confidence that nearly all the adversarial examples are outside the data manifold. To further provide support for this hypothesis we will perform an experiment where we ensure that these perturbations do not cause us to leave the manifold.

\subsection{What if we extend the data manifold to include the adversarial images?}
If adversarial attacks must go off manifold to succeed, then we should be able to defend against them by ensuring these perturbations are on manifold. 
An easy way to include adversarially attacked images in the data manifold is by using Gaussian data augmentation during training. Intuitively, this inflates the manifold like a balloon around the clean examples. Please note that such an idea was proposed in \cite{zantedeschi2017efficient} and also used in \cite{DBLP:conf/icml/GilmerFCC19}, but they did not tie it explicitly to the manifold hypothesis. Another key difference is that because our dataset is synthetic, we can be confident of avoiding confounding factors such as spurious correlations.

We train with Gaussian noise added to the pixels with standard deviation $\sigma=0.08$. The expected $\ell^2$ norm of the Gaussian noise is $4.8$. Thus, adversarial perturbations with
a norm of $\epsilon=0.15$ lie well within the data manifold. We see in Fig.~\ref{fig:pixel_eval} that for $\epsilon=0.15$ the adversarial accuracy was 92.84\% (compared to 12.86\% without Gaussian augmentation). In this experiment 92\% of successful adversarial attacks could be explained by the off manifold hypothesis.  

In addition we see in  Fig.~\ref{fig:distance_latent_noise} that the distance distribution for the adversarial examples is similar to that of mistakes.
Finally, Fig.~\ref{fig:adv_nois_self} illustrates that the adversarial errors are made on images that are also qualitatively more difficult. 

Please note (Fig.~\ref{fig:pixel_eval}) that the PGD attack is much more effective than a Gaussian noise attack at the same $\epsilon$. This is despite both samples having equal likelihood. This effect can partially be explained by the Gaussian Isoperimetric Inequality as discussed by Gilmer et al. \cite{DBLP:conf/icml/GilmerFCC19}. Their theorem states roughly that if Gaussian noise causes frequent errors at a certain $\epsilon$, then there exist
errors for a smaller $\epsilon$ that are unlikely to be revealed by random perturbations.  

\begin{figure}[t]
     \centering
    \begin{subfigure}{0.32\columnwidth}
     \includegraphics[trim=0 0 220 0,clip,width=\columnwidth,clip]{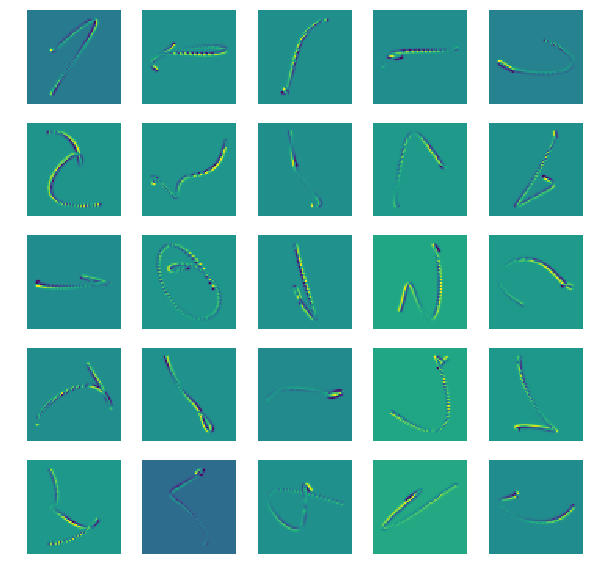}
     \caption{}\label{fig:attacks_latent}
     \end{subfigure}
     \begin{subfigure}{0.32\columnwidth}
     \includegraphics[trim=0 0 220 0,clip,width=\columnwidth,clip]{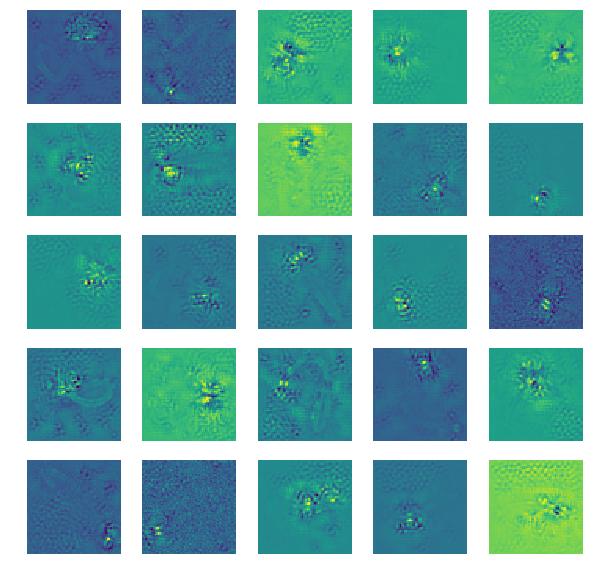}
     \caption{}\label{fig:attacks_std}
     \end{subfigure}
     \begin{subfigure}{0.32\columnwidth}
     \includegraphics[trim=0 0 220 0,clip,width=\columnwidth,clip]{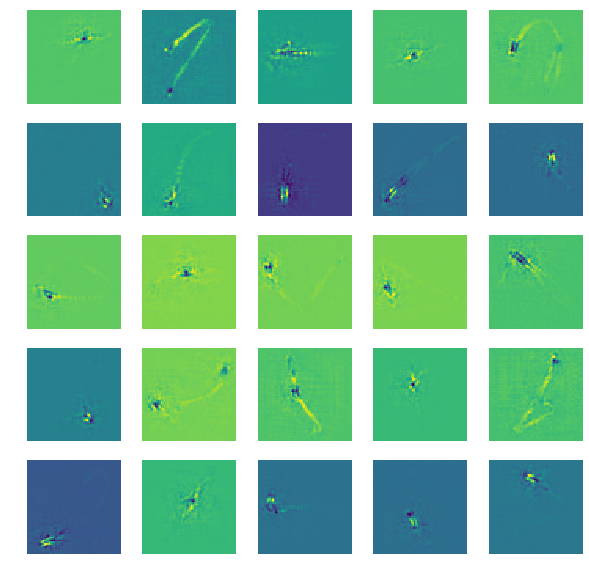}
     \caption{}\label{fig:attacks_noise}
     \end{subfigure}
     \caption{Comparison between adversarial attacks \subref{fig:attacks_latent} in latent space for a standard model, \subref{fig:attacks_std} in pixel space for a standard model and \subref{fig:attacks_noise} in pixel space for a model trained with pixel noise $\sigma=0.08$. The visualized attacks correspond to the samples in the last three panels of Fig.~\ref{fig:errors}.}
     \label{fig:attacks}
 \end{figure}

Furthermore, we see in Fig.~\ref{fig:attacks_noise} that the attack vector is highly focused on a semantically meaningful region, which was not the case for traditional adversarial examples. By using the differentiable generative process, we decomposed the attack into an on manifold component and an off manifold component. Even though the attack appears to be semantically meaningful, most of the power came from the off manifold contribution. 

\subsection{How applicable are these results to typical natural image datasets?}
In the experiments we have demonstrated that training with Gaussian noise can be an excellent defense against adversarial examples. However, in these experiments this defense was much more effective than when Gilmer et al. \cite{DBLP:conf/icml/GilmerFCC19} evaluated a similar approach on CIFAR-10 and ImageNet. Therefore we should think about how well this could translate to real world applications.

\begin{figure}
    \centering
    \begin{subfigure}{0.45\columnwidth}
    \includegraphics[width=\textwidth]{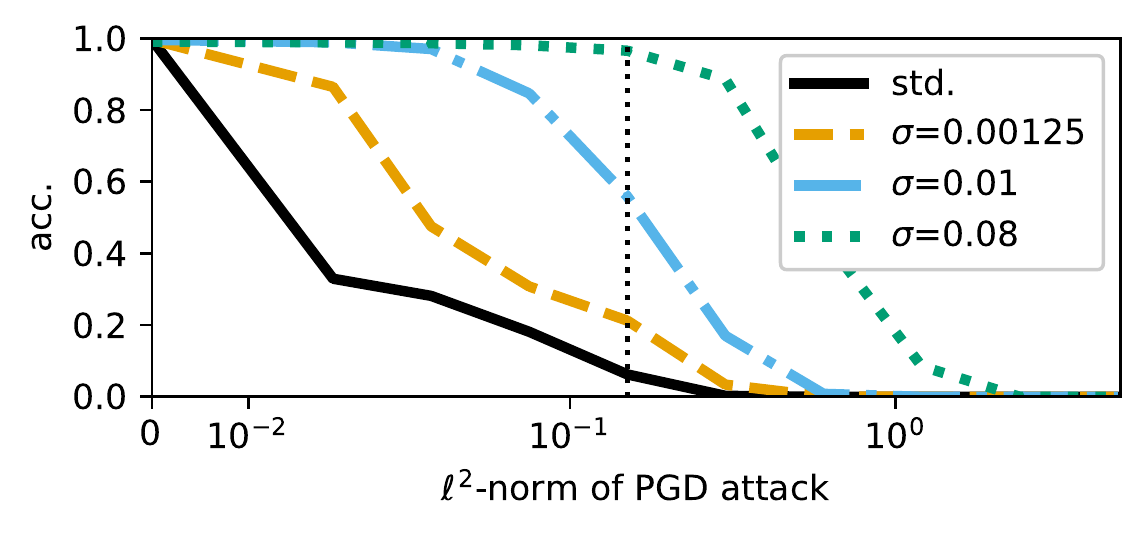}
    \caption{Pixel space}\label{fig:large_pixel}

    \end{subfigure}
    \begin{subfigure}{0.45\columnwidth}
    \includegraphics[width=\textwidth]{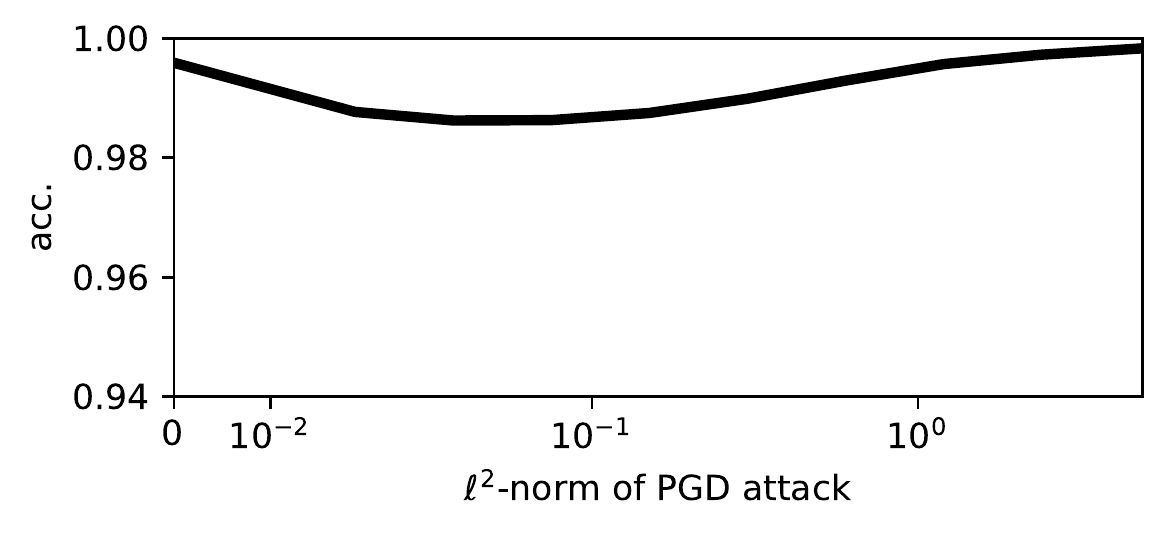}
    \caption{Latent space}\label{fig:large_latent}

    \end{subfigure}
    \caption{Comparison between adversarial attacks in pixel and latent space: high resolution variation. When trained without noise the model is more accurate but less robust in pixel space. Training with noise increases yields increased robustness. Adversarial examples in latent space are incredibly rare.}\label{fig:eval_latent_hires}
\end{figure}

First, it is important to note that training with Gaussian noise as a defense becomes more difficult as the number of dimensions increases. To demonstrate this we increase the image resolution from $60\times60$ to $120\times120$. The test accuracy increases to 99.59\% and for the most effective on manifold attack the accuracy only drops to 98.62\% (see Fig.~\ref{fig:large_latent}). But in pixel space, adversarial vulnerability increases. When we train with Gaussian noise, robustness re-appears and in fact improves for attacks of the same magnitude: $\ell^2=0.15$ results in an accuracy of 96.50\%. But since we have quadrupled the number of pixels, we should
double the attack magnitude to get the same level of perturbation per pixel. At a perturbation of $\ell^2=0.30$ the accuracy drops to 88.52\%. By this standard, the pixel-space robustness is actually a bit lower even with a Gaussian noise defense (Fig.~\ref{fig:large_pixel}). Since the volume around the manifold increases exponentially as we increase the magnitude of the attack, making sure that we define the classifier properly can require even more data. For that reason we do not expect it to scale indefinitely.

Second, we tried to uncover the fundamental problem causing adversarial examples in an idealized scenario. The amount of data we required to get a near perfect classifier is multiple times that what is available in ImageNet. In a dataset like ImageNet \cite{russakovsky2015imagenet}, the correct label for an image is not always straightforward and there are often multiple valid options. This leads to an ambiguous learning problem that can also result in a sub-optimal classifier. Based on our results, an ImageNet-like dataset with more and cleaner data would still suffer from adversarial examples, although Gaussian noise training would become a somewhat more effective defense.

\section{Discussion and Relation to Prior Work}
Previous work has invested a lot of effort in making neural networks more robust to adversarial examples \cite{roth2019odds,cisse2017parseval,song2017pixeldefend,samangouei2018defense,buckman2018thermometer} and in attacking them \cite{athalye2018obfuscated,DBLP:conf/iclr/MadryMSTV18,goodfellow2014explaining,carlini2017adversarial,DBLP:conf/iclr/KurakinGB17a,dong2019evading,li2019nattack}. Despite this, there is still an ongoing debate about the cause of adversarial examples. 

To better understand what causes adversarial examples we conducted the study on clean and well controlled data. We tested the hypothesis that adversarial examples are much harder to find on the data manifold. This hypothesis was used in \cite{song2017pixeldefend, samangouei2018defense} to construct adversarial defenses with generative models. In practice this defense was defeated \cite{athalye2018obfuscated}. However, our experiments indicate that the off manifold hypothesis has validity. The key issue with these defenses is therefore not conceptual but practical: learning a high quality data manifold is incredibly difficult.

Both Gilmer et al. \cite{DBLP:conf/iclr/GilmerMFSRWG18} and Stutz et al. \cite{Stutz_2019_CVPR} observed adversarial errors on the data manifold. However, there are important differences to our work.
The adversarial spheres dataset of Gilmer et al. does not have the notion of an imperceptible perturbation since the high dimensional samples do not represent sensory data such as images or sounds.
Stutz et al., in addition to studying a synthetic toy problem with a well-understood data manifold, approximated data manifolds of conventional datasets using generative models. They predicted that when test error decreases, on-manifold adversarial error decreases too. Both papers made the assumption that the classes are on separate, disconnected manifolds, an assumption which is often violated even on data such as MNIST where a 7 and a 1 can be indistinguishable. By making this assumption they excluded the decision boundary from latent space. Consequently they were not able to measure the
distance in latent space to the decision boundary as we were, and could not even attempt to assess whether a sample was more objectively
difficult. We have shown that when data is sufficiently available, on manifold errors are not only limited but also made mainly on objectively difficult samples, i.e., excusable.

Fawzi et al. \cite{fawzi2018adversarial}
provide theoretical evidence that non-zero test error entails adversarial examples, since some correctly classified samples will be close to generalization errors.  
We have shown that most generalization errors are close to the true decision boundary. When we combine this result with theirs, we can understand why the adversarial errors on manifold also are closer to the true decision boundary than typical samples but farther than typical errors.

Gilmer et al. \cite{DBLP:conf/icml/GilmerFCC19} performed experiments and provided evidence that Gaussian noise augmentation on the pixels can improve adversarial robustness. In their results a large fraction of samples can still be adversarially attacked. From their paper it is unclear why some examples became robust and others did not, which we studied specifically. 
Tram\`{e}r et al. \cite{DBLP:conf/icml/TramerBCPJ20} consider adversarial attacks where the ground truth label changes the but the model prediction does not. They rely on human labelers to judge the success of the attacks, whereas the labelling in this work is automated.

Finally, in our experiments there remain some samples that could be considered truly adversarial as they are mislabeled and relatively far from the decision boundary.  Ilyas et al. \cite{DBLP:conf/nips/IlyasSTETM19} make a strong argument that adversarial vulnerabilities are (in part) caused by non-robust features. Our experiments do not contradict this explicitly. While we observe a remarkable decrease in adversarial errors, we did not totally eliminate them.
Future work should try to understand whether the data can be adapted to modulate non-robust features explicitly such that this can be studied in isolation. 

In this paper we have used the Squiggles data to better understand the behavior of adversarial errors. While that is a key contribution in itself, we also believe that the dataset and its generative process have applications
far beyond adversarial examples.  We can easily extend the dataset to introduce subtle spurious correlations or shortcuts \cite{geirhos2020shortcut}
by sampling intersecting and non-intersecting curves from different distributions.
Because the difference between the distributions is so subtle, understanding the shortcut would make
a great challenge for interpretability methods, and suppressing it a great problem for inductive bias
studies.

\bibliography{references}
\bibliographystyle{alpha}

\cleardoublepage
\appendix

\section{The Squiggles dataset}\label{app:squiggles}
The curves in the squiggles dataset are generated by a process illustrated in Fig.~\ref{fig:generative} in the main paper. 

We start with a latent representation $\mathbf{z}$. This representation parametrizes a function $\boldsymbol{\Theta}_{\mathbf{z}}(t)$ that is able to generate a curve. We use this function to compute 100 coordinates $\mathbf{c}$ on the curve by applying $\boldsymbol{\Theta}_{\mathbf{z}}(t)$ to fixed inputs $t$. 
The coordinates $\mathbf{c}$ are used to compute the label and are also transformed into an image $\mathbf{x}$. The only component that is not continuous and piecewise differentiable is the computation of the label. In the remainder of this section we will detail the components. 

\subsection{Latent to coordinates}
We propose two approaches, Taylor and SineNet, to generate the coordinates. The first approach is based on Taylor series. The second variation, SineNet, uses random sinusoidal functions. Visualisations for SineNet are shown in Fig.~\ref{fig:dataset}; visualisations of Taylor are available in the appendix (Fig.~\ref{fig:taylor_dataset}).

\subsubsection{Taylor}
The Taylor data has latent parameters 
\[\mathbf{z} = (\mathbf{a}, \mathbf{b})^{\mathsf{T}} \in \mathbb{R}^{2 \times n}.\] In our experiments $n=7$.
The values for $\mathbf{a}, \mathbf{b}$ are randomly drawn from the uniform distribution over $[-1, 1]^n$. 
They represent the derivatives in the Taylor expansion of a curve:
\begin{align*}
\boldsymbol{\Theta}_{\mathbf{a}, \mathbf{b}} \colon \mathbb{R} &\rightarrow \mathbb{R}^2 \\
\boldsymbol{\Theta}_{\mathbf{a}, \mathbf{b}}(t) &= \flexparen{\sum_{i=1}^{n} \frac{a_i}{i!} t^i,
                                                                 \sum_{i=1}^{n} \frac{b_i}{i!} t^i}
\end{align*}
Based on aesthetic considerations, we take the portion of the curve traced out by $-3 \le t \le 3$.

\subsubsection{SineNet}

For the SineNet curves the latent space
\[
\mathbf{z} = (\mathbf{a}, \mathbf{b}, \boldsymbol{\omega}, \boldsymbol{\phi})^{\mathsf{T}} \in \mathbb{R}^n \times \mathbb{R}^n \times \mathbb{R}^n \times [0, 2\pi]^n
\] is much larger with $n=300$ leading to a 1200 dimensional latent code. 
\begin{equation}\label{eqn:sine_net}
\begin{split}
    \boldsymbol{\Theta}_{\mathbf{a}, \mathbf{b}, \boldsymbol{\omega}, \boldsymbol{\phi}} \colon \mathbb{R} &\rightarrow \mathbb{R}^2 \\
    \boldsymbol{\Theta}_{\mathbf{a}, \mathbf{b}, \boldsymbol{\omega}, \boldsymbol{\phi}}(t) &= \sum_{i=1}^n (a_i, b_i) \cdot \sin(\omega_i t + \phi_i)
\end{split}
\end{equation}
Here, $\mathbf{a}, \mathbf{b}$, which represent the amplitude of the sinusoids, are drawn from a standard normal distribution, as is the frequency $\boldsymbol{\omega}$. The phase $\boldsymbol{\phi}$ is drawn from a uniform distribution. Based on aesthetic considerations, we take the portion of the curve traced out by
$-2 \leq t \leq 2$.

Not coincidentally, \eqref{eqn:sine_net} is an equation for a neural net with one input neuron, two output neurons, and
a single hidden layer with sinusoidal activations. Hence the name SineNet.

Qualitatively speaking, the curves generated by SineNet appear to be more curvy than those of the Taylor variation. Therefore, we can see that these distributions are different, despite only differing in how the points are generated. 

\subsubsection{Curve to coordinates}\label{curve_to_planar}

Given $\mathbf{z}$ we can compute the planar coordinates 
\begin{equation*}
\mathbf{c} = \begin{pmatrix*}[l]
    \boldsymbol{\Theta}_{\mathbf{z}}(-3) \\
    \boldsymbol{\Theta}_{\mathbf{z}}(-3 + \tfrac{2}{33})) \\
    \boldsymbol{\Theta}_{\mathbf{z}}(-3 + \tfrac{4}{33})) \\
    \mask{\boldsymbol{\Theta}}{\vdots} \\
    \boldsymbol{\Theta}_{\mathbf{z}}(3)
\end{pmatrix*} \in \mathbb{R}^{100 \times 2}
\end{equation*}

[Note: these are the $t$-values for Taylor. For SineNet, the $t$-values
range over $[-2, 2]$ rather than $[-3, 3]$.]

We now have 100 points on the curve $\boldsymbol{\Theta}_{\mathbf{z}}$. From here on out, we consider the curve to be the
piecewise-linear curve connecting these 100 points sequentially.

\subsection{Coordinates to labels}

\begin{figure}
\centering
\includegraphics{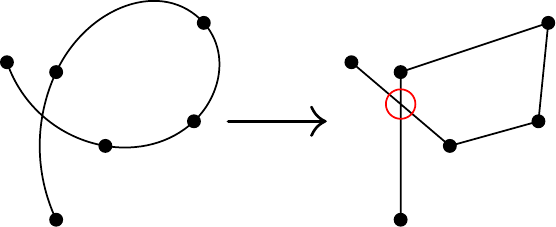}
\caption{To compute whether a curve is self-intersecting, it is first converted to a piecewise-linear path so
that standard algorithms can more easily be applied. Of course, 100 points trace out a path much more tightly than the
six points shown here.}\label{self_isecting_polygon}
\end{figure}

We already have a list of 100 planar points. Using the
\href{https://shapely.readthedocs.io/en/stable/manual.html}{Shapely package}
\cite{shapely2007} we evaluate whether the piecewise-linear curve that
connects these points sequentially is self-intersecting, and label the
example accordingly. See Figure \ref{self_isecting_polygon}. We do not attempt to account for curves that gain or lose a
self-intersection by such polygonization; we assume such curves would be difficult
to correctly categorize in any case since the visual representation depends only
on the 100 points.

The label distributions for Taylor and SineNet are given in Table~\ref{table:label_distributions}. Generally speaking self-intersecting curves are less common than simple curves.

\subsection{Coordinates to pixels}

To create a grayscale image from 100 $(x,y)$-points, we color each pixel according to its distance to the nearest point.
See below for the detailed steps. Note that each item represents a continuous, piecewise-differentiable
function of the previous output.

\begin{notation}
The symbols $c_{k,x}$ and $c_{k,y}$ denote the $x$- and $y$-coordinates, respectively, of the ordered
pair that is the $k^{\textrm{th}}$ row of $\mathbf{c}$, defined in \ref{curve_to_planar}. Similarly, $\mathbf{c}_{x},
\mathbf{c}_{y}$ denote the first and second columns of $\mathbf{c}$.

When an affine transformation is applied to the curve points $\mathbf{c}$, we will continue to use the symbol $\mathbf{c}$ to
denote the transformed curve. And likewise for its coordinates $c_{k,x}$ and $c_{k,y}$.

The matrix of pixel intensities is an
$n \times n$ matrix denoted by $\mathbf{X}$. Its coordinates are denoted $X_{ij}$ where $i, j \in \set{1,2\cmb\dotsc,n}$.
There is a closely related tensor $X_{ijk}$, denoting the contribution of the $k^{\textrm{th}}$ point of the curve to the
$(i,j)^{\textrm{th}}$ pixel. (Do not take the term \textit{contribution} too seriously: in max
pooling only one curve point actually ends up ``contributing" to any one pixel.)
\end{notation}

\begin{enumerate}[label=\arabic*.]
  \item Rescale the points (preserving the aspect ratio) so the maximum distance between any two points along the $x$ or $y$ directions
        is $0.8$.
        \begin{equation*}
            \mathbf{c} \mapsto \frac{0.8}{\max\flexset{\splitfrac{\max_k c_{k,x} - \min_k c_{k,x},}
                {\max_k c_{k,y} - \min_k c_{k,y}}}} \cdot \mathbf{c}
        \end{equation*}
  \item Translate the points to center the curve within the box $[0, 1] \times [0, 1]$.
        \begin{align*}
            c_{k,x} &\mapsto c_{k,x} + \frac{1 - \max_{\ell} c_{\ell,x} - \min_{\ell} c_{\ell,x}}{2}  \\
            c_{k,y} &\mapsto c_{k,y} + \frac{1 - \max_{\ell} c_{\ell,y} - \min_{\ell} c_{\ell,y}}{2}
        \end{align*}
    \item For the $k^{\textrm{th}}$ $(x,y)$-point on the curve, compute an image $\mathbf{X}_{\bullet\bullet k}$. In this image pixel $(i,j)$ 
        represents the distance from that pixel to the point $\mathbf{c}_k$, after
        applying a Gaussian RBF:
        \begin{multline*}
            X_{ijk} = \exp\Biggl(-\frac{1}{(1/n)^2} \cdot \biggl(\flexparen{\frac{i-1}{n-1} - c_{k,x}}^2 \\ + \flexparen{\frac{j-1}{n-1} - c_{k,y}}^2\biggr)\Biggr)
        \end{multline*}
        Thus, a point gets translated into a ``fuzzy point" to allow for differentiability and also provide anti-aliasing.

        The particular RBF parameters are chosen so the peak value is $1.0$ and 
        the inflection point is approximately one pixel-width
        away from the curve point.
    \item Take the maximum among all $k$ curve points to get the intensity based on the closest point:
        \begin{equation*}
            X_{ij} = \max_k X_{ijk}
        \end{equation*}
        We use maximum rather than mean or sum to get visual representations more similar to drawings.
        Mean or sum would produce a cloud of greater intensity
        in segments of closely spaced points (e.g. tight loops), making the rest of the curve look washed-out by comparison.
\end{enumerate}

\noindent\begin{minipage}{\linewidth}\setcaptiontype{figure}
\centering
\includegraphics{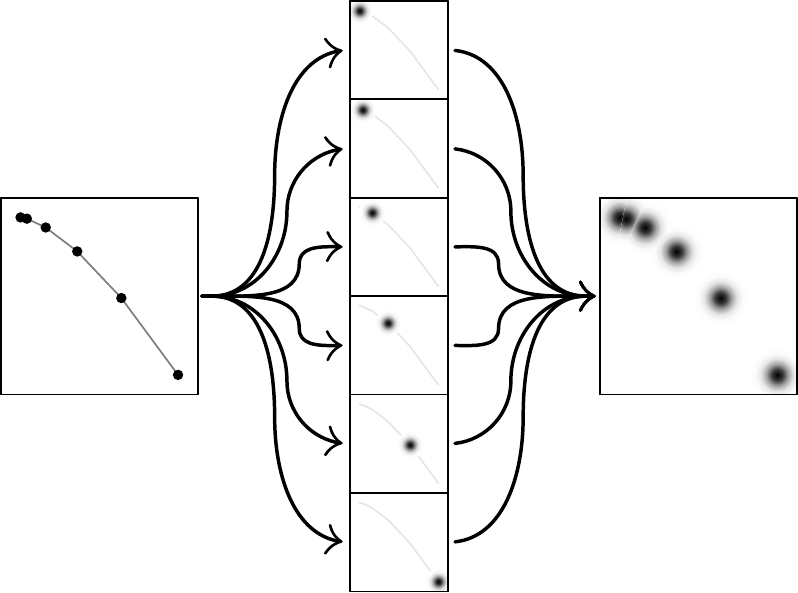}
\caption{To turn coordinates into an image, we first compute an image for each coordinate separately by drawing a Gaussian centered at its location. The final value of a pixel is its maximum value over all sub-images.}
\end{minipage}\vspace{.5cm}

\subsubsection{Pixel intensities away from the curve at 32-bit float precision}

Because the pixel intensity algorithm uses a Gaussian activation, the theoretical (infinite-precision) value of the
pixel intensity is nonzero over the entire image, even at pixels nowhere near the actual curve. This is potentially
desirable for better latent-space derivatives, since a pixel far away from the curve can still attract or repel the
nearest point on the curve if that pixel has sufficiently high salience. However, this feature is undesirable in that
it could allow models to learn features based on subtle shadings that are invisible to the human eye.

When one leaves the land of infinite-precision reals and considers the 32-bit floats that are used for the actual
implementation, it turns out that neither of these features is as significant as it first appears. According to the
IEEE standard \cite{IEEE_float_standard}, the smallest positive\footnote{If subnormal numbers are disallowed, 
this becomes $2^{-126}$, which gives a 
radius of $9.38 / n$.} number that can be represented in 32-bit floating arithmetic is
$2^{-149}$. Consequently, any pixel with a theoretical intensity less than $2^{-150}$ has a computed intensity of zero.
Solving for the distance $d$ from the nearest point gives
\begin{align*}
    \exp(-n^2 \cdot d^2) &< 2^{-150} \\
    d &> \frac{1}{n}\sqrt{150 \log 2} \approx 10.20 / n
\end{align*}
Thus, nonzero values are constrained to a distance of about 10 pixel-widths from the curve. (Recall that $n$ is the number of pixels along
one side of a $1\times1$ square, so $1/n$ is approximately the width of a pixel.)

 \begin{figure*}
     \centering
     \begin{subfigure}[t]{0.49\textwidth}
		\centering		
		\includegraphics[width=\textwidth]{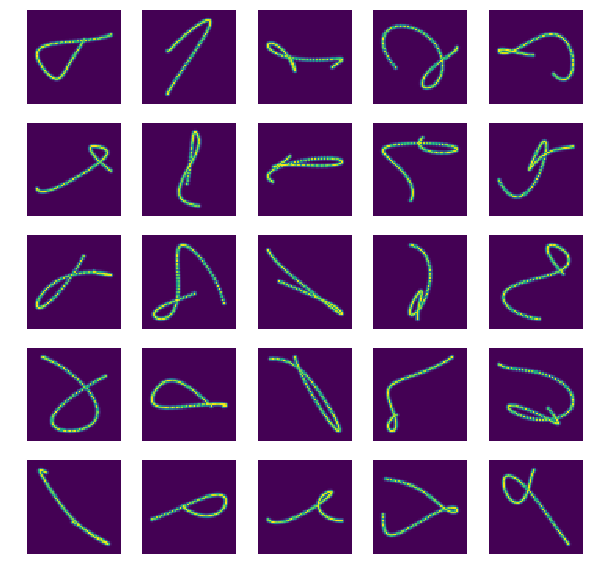}
		\caption{self-intersecting}\label{fig:sinenet_rng_self}		
	\end{subfigure}
     \begin{subfigure}[t]{0.49\textwidth}
		\centering		
		\includegraphics[width=\textwidth]{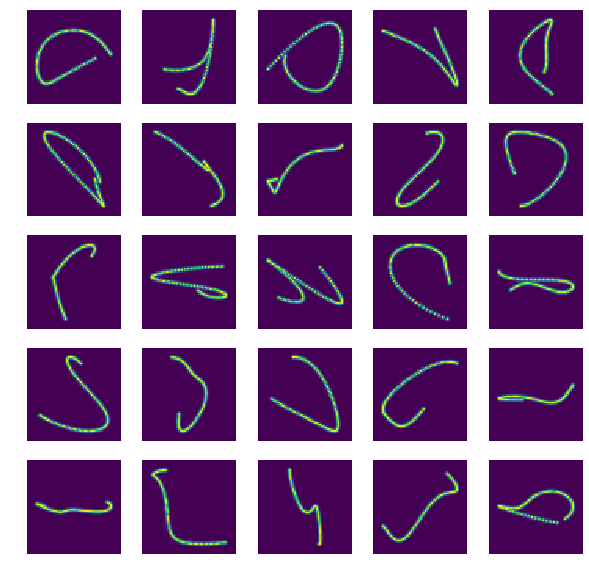}
		\caption{not-intersecting}\label{fig:sinenet_rng_not}		
	\end{subfigure}
	\caption{\textbf{SineNet random examples}: corresponds to Fig.~\ref{fig:dataset} in the paper.}
 \end{figure*}

 \begin{figure*}
     \centering
     \begin{subfigure}[t]{0.49\textwidth}
		\centering		
		\includegraphics[width=\textwidth]{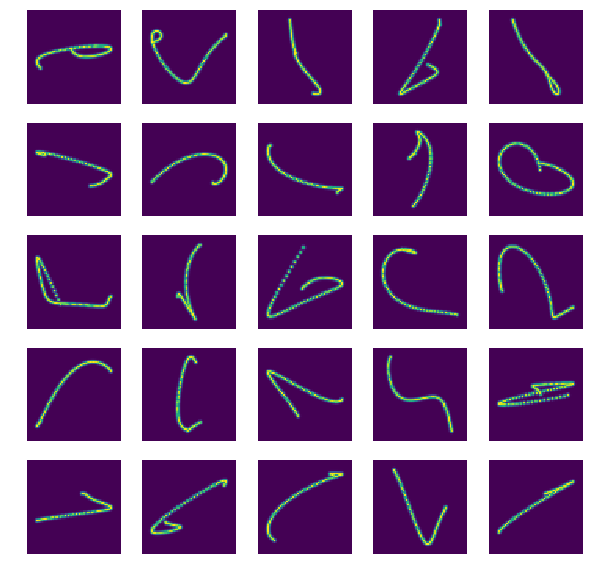}
		\caption{self-intersecting}\label{fig:sinenet_mistake_self}		
	\end{subfigure}
     \begin{subfigure}[t]{0.49\textwidth}
		\centering		
		\includegraphics[width=\textwidth]{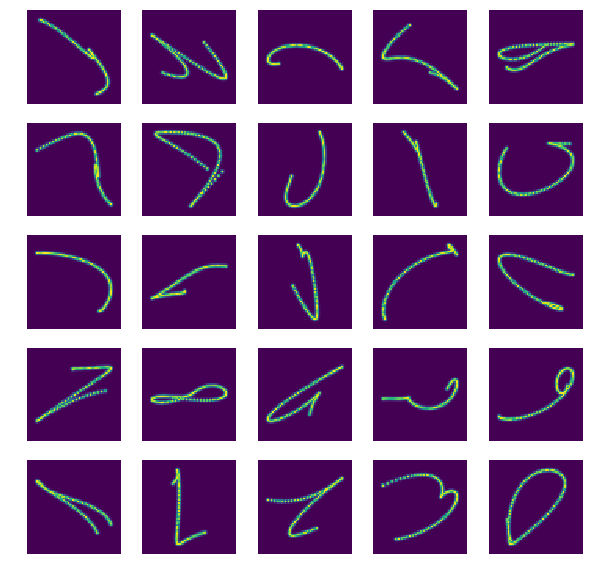}
		\caption{not-intersecting}\label{fig:sinenet_mistake_not}		
	\end{subfigure}
	\caption{\textbf{SineNet mistakes}. The images are grouped by ground truth label. Corresponds to Fig.~\ref{fig:mistake_self} in the paper.}
 \end{figure*}
 \begin{figure*}

    \begin{subfigure}[t]{0.49\textwidth}
		\centering		
		\includegraphics[width=\textwidth]{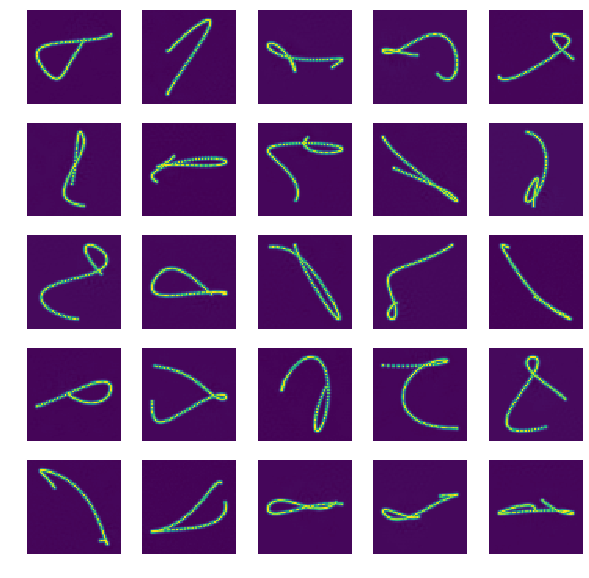}
		\caption{self-intersecting}\label{fig:sinenet_adv_self}		
	\end{subfigure}
     \begin{subfigure}[t]{0.49\textwidth}
		\centering		
		\includegraphics[width=\textwidth]{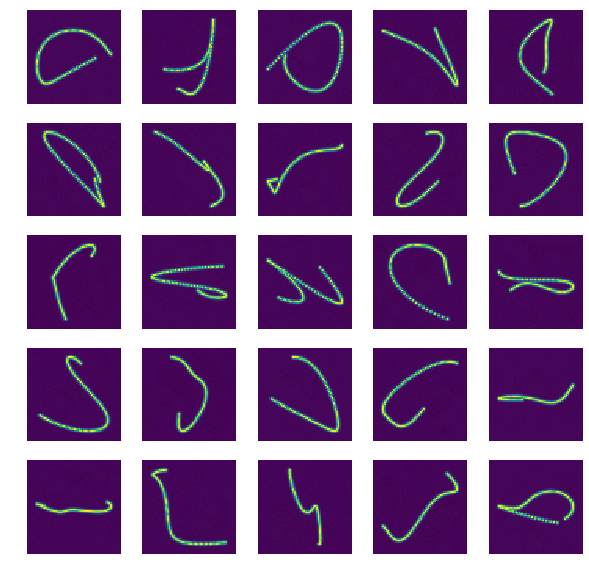}
		\caption{not-intersecting}\label{fig:sinenet_adv_not}		
	\end{subfigure}
	\caption{\textbf{SineNet adversarial errors (model trained without noise $\sigma=0$)}. The images are grouped by ground truth label. Corresponds to Fig.~\ref{fig:adv_self} in the paper.} \end{figure*}
 \begin{figure*}

    \begin{subfigure}[t]{0.49\textwidth}
		\centering		
		\includegraphics[width=\textwidth]{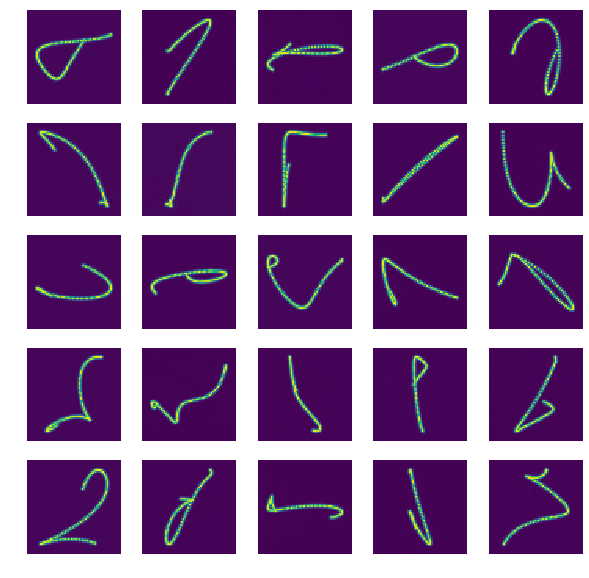}
		\caption{self-intersecting}\label{fig:sinenet_noise_self}		
	\end{subfigure}
     \begin{subfigure}[t]{0.49\textwidth}
		\centering		
		\includegraphics[width=\textwidth]{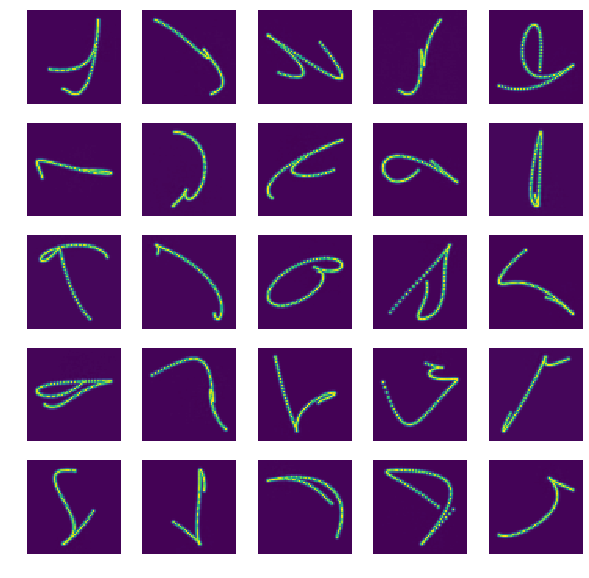}
		\caption{not intersecting}\label{fig:sinenet_noise_not}		
	\end{subfigure}
	\caption{\textbf{SineNet adversarial errors (model trained with noise $\sigma=0.08$)}. The images are grouped by ground truth label. Corresponds to Fig.~\ref{fig:adv_nois_self} in the paper.} \end{figure*}
	
 \begin{figure*}
    \begin{subfigure}[t]{0.49\textwidth}
		\centering		
		\includegraphics[width=\textwidth]{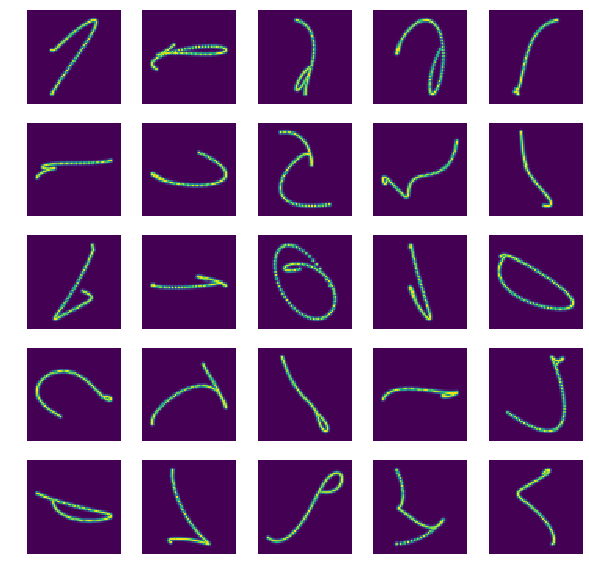}
		\caption{self-intersecting}\label{fig:sinenet_latent_self}		
	\end{subfigure}
     \begin{subfigure}[t]{0.49\textwidth}
		\centering		
		\includegraphics[width=\textwidth]{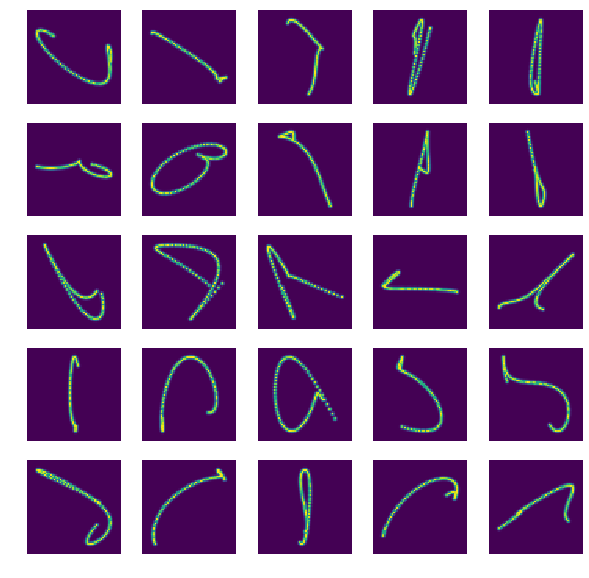}
		\caption{not intersecting}\label{fig:sinenet_latent_not}		
	\end{subfigure}	
	\caption{\textbf{SineNet adversarial errors in latent space (model trained without noise $\sigma=0$)}. The images are grouped by ground truth label. Corresponds to Fig.~\ref{fig:adv_self} in the paper.}\end{figure*}

 \begin{figure*}
     \centering
     \begin{subfigure}[t]{0.49\textwidth}
		\centering		
		\includegraphics[width=\textwidth]{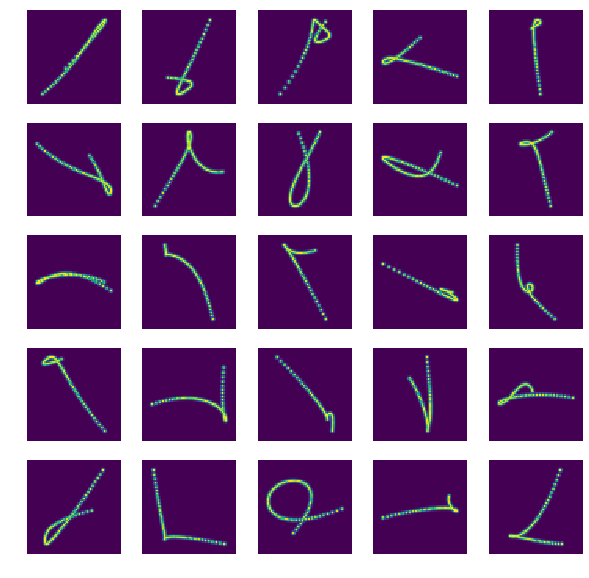}
		\caption{self-intersecting}\label{fig:taylor_rng_self}		
	\end{subfigure}
     \begin{subfigure}[t]{0.49\textwidth}
		\centering		
		\includegraphics[width=\textwidth]{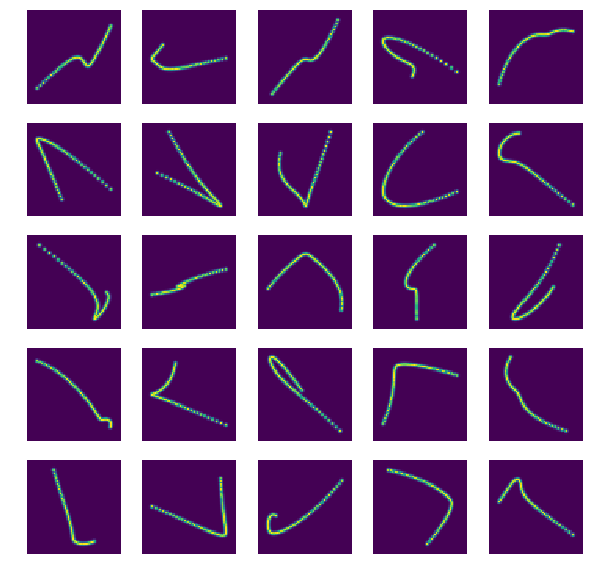}
		\caption{not-intersecting}\label{fig:taylor_rnd_not}		
	\end{subfigure}
	\caption{\textbf{Taylor random examples}: corresponds to Fig.~\ref{fig:dataset} in the paper.\label{fig:taylor_dataset}}
 \end{figure*}

 \begin{figure*}
     \centering
     \begin{subfigure}[t]{0.49\textwidth}
		\centering		
		\includegraphics[width=\textwidth]{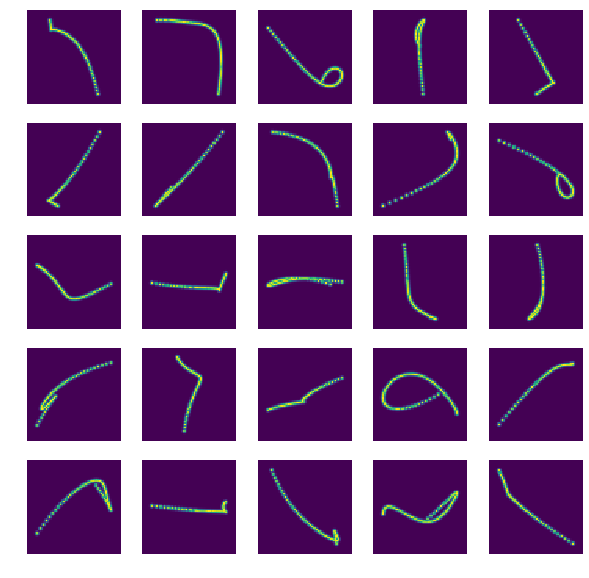}
		\caption{self-intersecting}\label{fig:taylor_mistake_self}		
	\end{subfigure}
     \begin{subfigure}[t]{0.49\textwidth}
		\centering		
		\includegraphics[width=\textwidth]{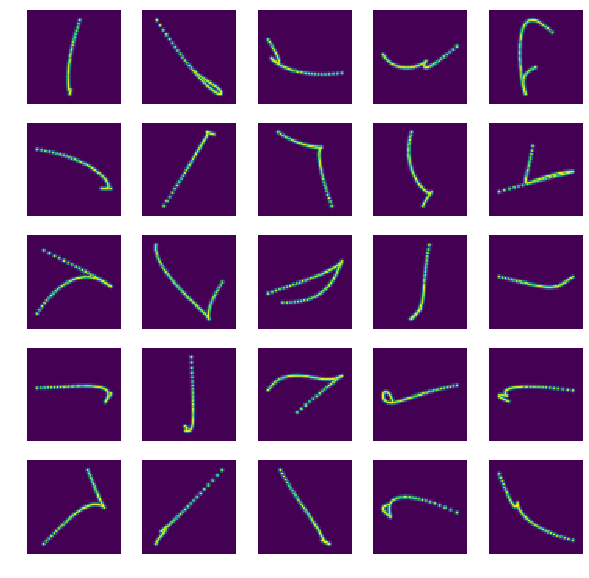}
		\caption{not-intersecting}\label{fig:taylor_mistake_not}		
	\end{subfigure}
	\caption{\textbf{Taylor mistakes}. The images are grouped by ground truth label. Corresponds to Fig.~\ref{fig:mistake_self} in the paper.}
 \end{figure*}
 \begin{figure*}

    \begin{subfigure}[t]{0.49\textwidth}
		\centering		
		\includegraphics[width=\textwidth]{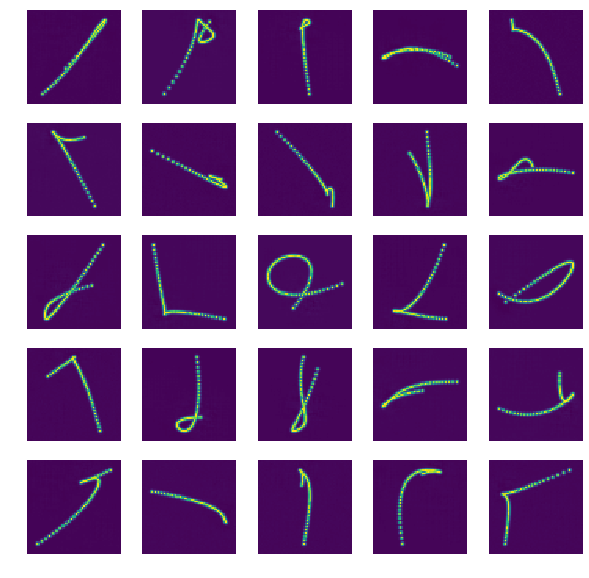}
		\caption{self-intersecting}\label{fig:taylor_adv_self}		
	\end{subfigure}
     \begin{subfigure}[t]{0.49\textwidth}
		\centering		
		\includegraphics[width=\textwidth]{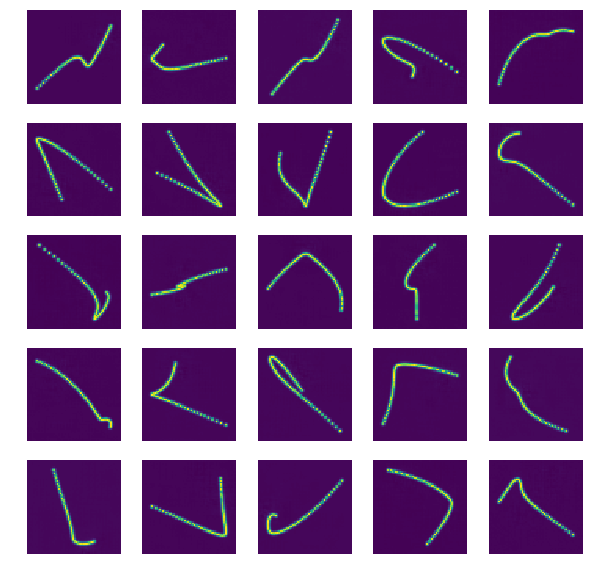}
		\caption{not-intersecting}\label{fig:taylor_adv_not}		
	\end{subfigure}
	\caption{\textbf{Taylor adversarial errors (model trained without noise $\sigma=0$)}. The images are grouped by ground truth label. Corresponds to Fig.~\ref{fig:adv_self} in the paper.} \end{figure*}
 \begin{figure*}

    \begin{subfigure}[t]{0.49\textwidth}
		\centering		
		\includegraphics[width=\textwidth]{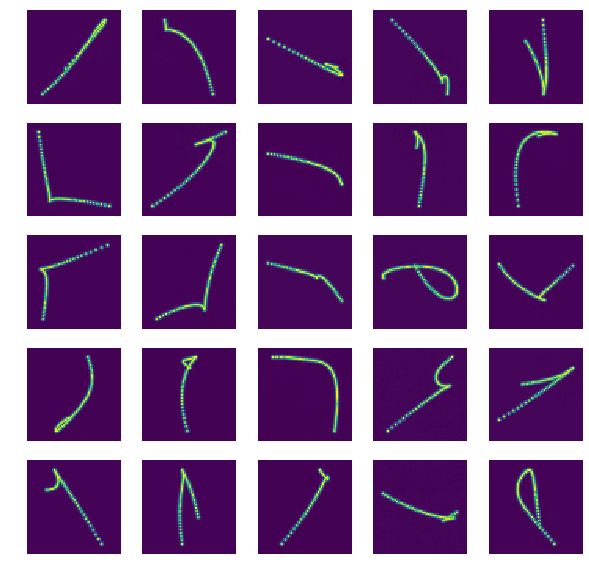}
		\caption{self-intersecting}\label{fig:taylor_noise_self}		
	\end{subfigure}
     \begin{subfigure}[t]{0.49\textwidth}
		\centering		
		\includegraphics[width=\textwidth]{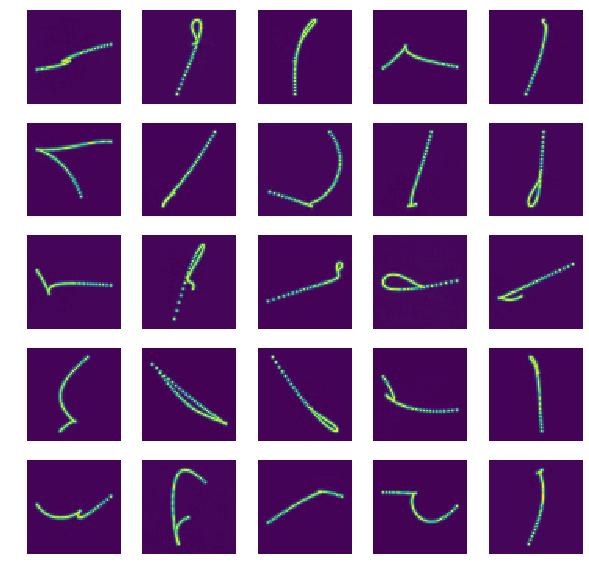}
		\caption{not intersecting}\label{fig:taylor_noise_not}		
	\end{subfigure}
	\caption{\textbf{Taylor adversarial errors (model trained with noise $\sigma=0.08$)}. The images are grouped by ground truth label. Corresponds to Fig.~\ref{fig:adv_nois_self} in the paper.} \end{figure*}
	
 \begin{figure*}
    \begin{subfigure}[t]{0.49\textwidth}
		\centering		
		\includegraphics[width=\textwidth]{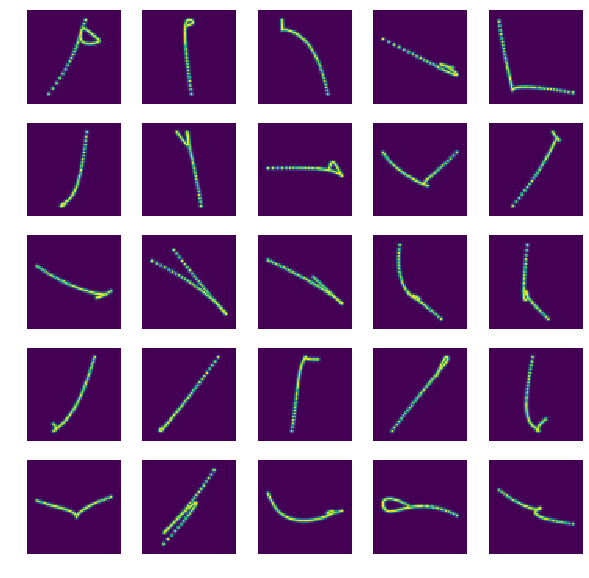}
		\caption{self-intersecting}\label{fig:taylor_latent_self}		
	\end{subfigure}
     \begin{subfigure}[t]{0.49\textwidth}
		\centering		
		\includegraphics[width=\textwidth]{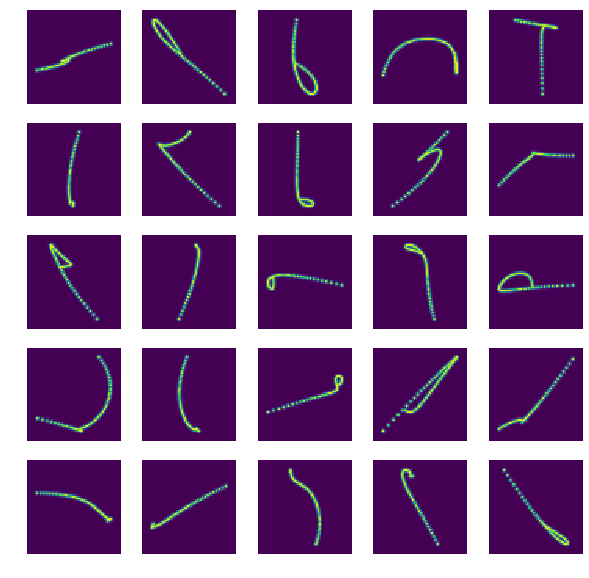}
		\caption{not intersecting}\label{fig:taylor_latent_not}		
	\end{subfigure}	
	\caption{\textbf{Taylor adversarial errors in latent space (model trained without noise $\sigma=0$)}. The images are grouped by ground truth label. Corresponds to Fig.~\ref{fig:adv_self} in the paper.}\end{figure*}

\begin{figure*}
\centering
\includegraphics[width=0.4\textwidth]{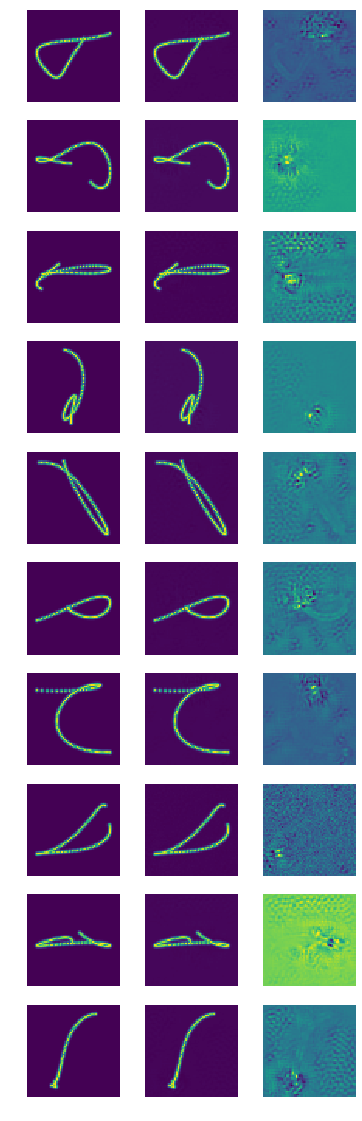}
\caption{Adversarial errors after training without noise $\sigma=0$. The attack has a norm of 0.15. From left to right, original image, attacked image, attack vector. When we do n ot train with noise, the attack has a focus point on semantically meaningful regions, but also has a wide presence by generating a checkerboard pattern.}\label{fig:sup:attack_reg}
\end{figure*}
\begin{figure*}
\centering
\includegraphics[width=0.4\textwidth]{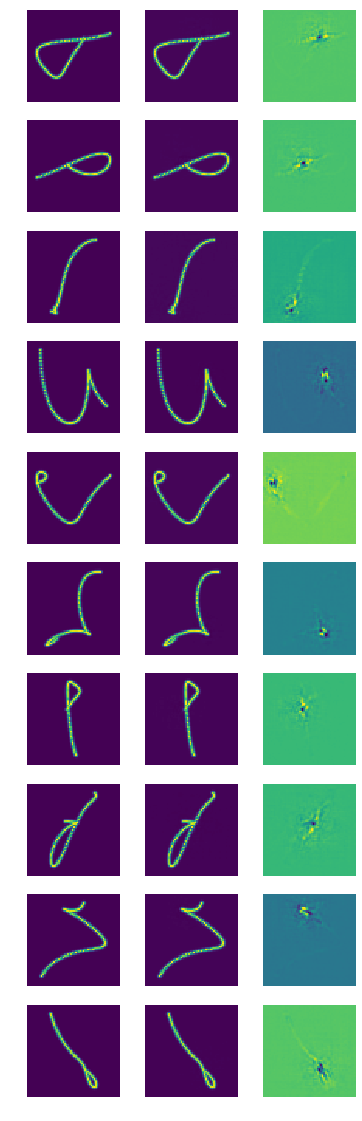}
\caption{Adversarial errors after training on noise $\sigma=0.8$. The attack has a norm of 0.15. From left to right, original image, attacked image, attack vector. Unlike for the model trained without noise. This attack does not have the widespread checkerboard pattern. }\label{fig:sup:attack_noise}
\end{figure*}

\section{Taylor visualizations}

We present visualizations of the Taylor dataset, which were omitted from the
paper itself because of space considerations.

\section{Non-self-intersecting visualizations}
Certain visualizations in the paper itself were presented only for self-intersecting
curves. We present non-self-intersecting counterparts to these visualizations.

\section{Visualizing the attack and the images}
In the paper we visualised the attacks for a certain set of errors. Unfortunately, we were unable to put them side by side due to space constraints. Figures where they are side by side are included in Fig. \ref{fig:sup:attack_reg} and Fig. \ref{fig:sup:attack_noise}.

\section{Visualizing the intersections on the mistaken images}
\begin{figure}
    \centering
    \includegraphics[width=\columnwidth]{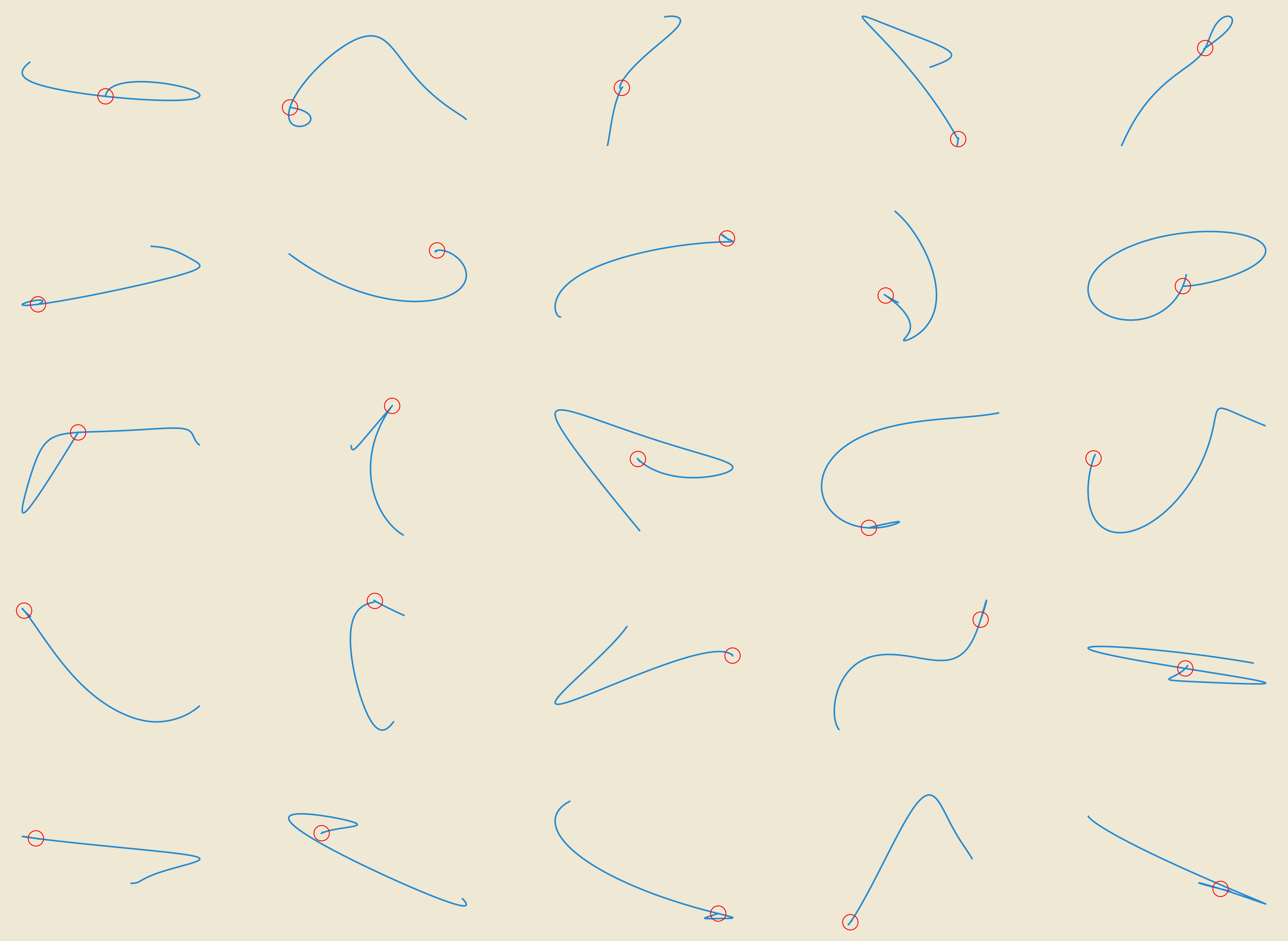}
    \caption{The computed intersection points for the curves of \ref{fig:mistake_self}, which the model mistakenly (but understandably)
    believed were not self intersecting. If you are viewing this digitally, consider zooming in to view it at around 5x magnification.}
    \label{fig:show_intersections}
\end{figure}

The classification errors on the self intersections are not obvious. Therefore we include an enlarged visualization with the location of the self intersection highlighted in red. This is shown in Fig. \ref{fig:show_intersections}.

\end{document}